\documentclass[12pt]{article}
\usepackage{amsmath,amsthm,amsbsy,amsfonts}
\usepackage{bm,graphicx,psfrag,epsf}
\usepackage{enumerate}
\usepackage[round, authoryear]{natbib}
\usepackage{url} 
\usepackage{bbm}
\usepackage{ifpdf}

\usepackage{setspace}
\usepackage[usenames]{color}
\RequirePackage[OT1]{fontenc}
\usepackage{algorithm}
\usepackage{algpseudocode}

\usepackage{mathtools}

\usepackage{paralist}

\usepackage{amsbsy}

\usepackage[mathscr]{eucal}

\usepackage{bm}

\usepackage{xspace}

\usepackage{subfig}
\usepackage{multirow}

\mathchardef\mhyphen="2D

\ifpdf
\usepackage[colorlinks,urlcolor=blue,citecolor=blue,linkcolor=blue]{hyperref}
\else
\newcommand{\href}[2]{{#2}}
\usepackage{url}
\fi

\ifpdf
\newcommand{\Sec}[1]{\hyperref[sec:#1]{Section~\ref*{sec:#1}}} 
\newcommand{\App}[1]{\hyperref[sec:#1]{Appendix~\ref*{sec:#1}}} 
\newcommand{\Supp}[1]{\hyperref[sec:#1]{Supplement~\ref*{sec:#1}}} 
\newcommand{\Eqn}[1]{\hyperref[eq:#1]{{\rm (\ref*{eq:#1})}}} 
\newcommand{\Part}[1]{\hyperref[part:#1]{(\ref*{part:#1})}} 
\newcommand{\Fig}[1]{\hyperref[fig:#1]{Figure~\ref*{fig:#1}}} 
\newcommand{\Tab}[1]{\hyperref[tab:#1]{Table~\ref*{tab:#1}}} 
\newcommand{\Thm}[1]{\hyperref[thm:#1]{Theorem~\ref*{thm:#1}}} 
\newcommand{\Lem}[1]{\hyperref[lem:#1]{Lemma~\ref*{lem:#1}}} 
\newcommand{\Prop}[1]{\hyperref[prop:#1]{Proposition~\ref*{prop:#1}}} 
\newcommand{\Cor}[1]{\hyperref[cor:#1]{Corollary~\ref*{cor:#1}}} 
\newcommand{\Def}[1]{\hyperref[def:#1]{Definition~\ref*{def:#1}}} 
\newcommand{\Alg}[1]{\hyperref[alg:#1]{Algorithm~\ref*{alg:#1}}} 
\newcommand{\Ex}[1]{\hyperref[ex:#1]{Example~\ref*{ex:#1}}} 
\newcommand{\As}[1]{\hyperref[as:#1]{Assumption~{\rm\ref*{as:#1}}}} 
\newcommand{\Reg}[1]{\hyperref[as:#1]{Condition~\ref*{reg:#1}}} 
\newcommand{\AlgLine}[2]{\hyperref[alg:#1]{line~\ref*{line:#2} of Algorithm~\ref*{alg:#1}}}
\newcommand{\AlgLines}[3]{\hyperref[alg:#1]{lines~\ref*{line:#2}--\ref*{line:#3} of Algorithm~\ref*{alg:#1}}}
\else
\newcommand{\Sec}[1]{{Section~\ref{sec:#1}}} 
\newcommand{\App}[1]{{Appendix~\ref{sec:#1}}} 
\newcommand{\Supp}[1]{{Supplement~\ref{sec:#1}}} 
\newcommand{\Eqn}[1]{{(\ref{eq:#1})}} 
\newcommand{\Part}[1]{{(\ref{part:#1})}} 
\newcommand{\Fig}[1]{{Figure~\ref{fig:#1}}} 
\newcommand{\Tab}[1]{{Table~\ref{tab:#1}}} 
\newcommand{\Thm}[1]{{Theorem~\ref{thm:#1}}} 
\newcommand{\Lem}[1]{{Lemma~\ref{lem:#1}}} 
\newcommand{\Prop}[1]{{Proposition~\ref{prop:#1}}} 
\newcommand{\Cor}[1]{{Corollary~\ref{cor:#1}}} 
\newcommand{\Def}[1]{{Definition~\ref{def:#1}}} 
\newcommand{\Alg}[1]{{Algorithm~\ref{alg:#1}}} 
\newcommand{\Ex}[1]{{Example~\ref{ex:#1}}} 
\newcommand{\Reg}[1]{{R~\ref*{reg:#1}}} 
\fi

\newcommand{\Real}{\mathbb{R}}

\newcommand{\Tra}{^{\sf T}} 

\newcommand{\V}[1]{{\bm{\mathbf{\MakeLowercase{#1}}}}} 
\newcommand{\VE}[2]{\MakeLowercase{#1}_{#2}} 

\newcommand{\M}[1]{{\bm{\mathbf{\MakeUppercase{#1}}}}} 
\newcommand{\ME}[2]{\MakeLowercase{#1}_{#2}} 

\newcommand{\Mtilde}[1]{{\bm{\tilde \mathbf{\MakeUppercase{#1}}}}} 

\newcommand{\amp}{\mathop{\:\:\,}\nolimits}
\newcommand{\ampb}{\mathop{\:\,}\nolimits}

\definecolor{blue}{rgb}{0.2,0.5,0.7}
\definecolor{green}{rgb}{0.3,0.68,0.29}
\definecolor{purple}{rgb}{0.6,0.31,0.64}

\newtheorem{proposition}{Proposition}[section]
\newtheorem{corollary}{Corollary}[section]

\newcommand{\blind}{1}

\begin{document}

\def\spacingset#1{\renewcommand{\baselinestretch}%
{#1}\small\normalsize} \spacingset{1}

\if1\blind
{

  \title{\bf A Majorization-Minimization Gauss-Newton Method for 1-Bit Matrix Completion}
  \author{Xiaoqian Liu\\
    Department of Statistics\\
    University of California, Riverside\\
    Xu Han  \\
    Tencent, Beijing\\
    Eric C.\@ Chi \\
    Department of Statistics,
    Rice University\\
    and\\
    Boaz Nadler\\
    Department of Computer Science and Applied Mathematics\\
Weizmann Institute of Science
    }
    \date{}
  \maketitle
} \fi


\if0\blind
{
  \bigskip
  \bigskip
  \bigskip
  \begin{center}
    {\LARGE\bf A Majorization-Minimization Gauss-Newton Method for 1-Bit Matrix Completion}
\end{center}
  \medskip
} \fi

\begin{abstract}
In 1-bit matrix completion, the aim is to estimate an underlying low-rank matrix from a partial set of binary observations. We propose a novel method for 1-bit matrix completion called Majorization-Minimization Gauss-Newton (\texttt{MMGN}). Our method is based on the
majorization-minimization principle, which converts the original optimization problem into a sequence of standard low-rank matrix completion problems.
%
%
 We solve each of these sub-problems by a factorization approach that explicitly enforces the assumed low-rank structure and then apply a Gauss-Newton method. 
 Using simulations and a real data example, we illustrate that 
 in comparison to existing 1-bit matrix completion methods, 
 \texttt{MMGN} outputs comparable if not more accurate estimates. 
 In addition, it is 
often significantly faster, and less sensitive to the spikiness of the underlying matrix. 
In comparison with three standard generic optimization approaches
that directly minimize the original objective, \texttt{MMGN} also exhibits a clear computational advantage, especially when the fraction of observed entries is small.
\end{abstract}

\noindent%
\vfill

\newpage
\spacingset{1.45} 

\section{Introduction}
\label{sec: intro}

In this work, we consider the following 1-bit matrix completion problem: 
Let
$\M{\Theta}^* \in \Real^{m\times n}$ be an unknown matrix of exact rank $r^*$.
The observed data are a subset $\Omega \subset [m]\times [n]$  and a binary matrix $\M{Y}$. At  
entries $(i,j)\notin\Omega$, $y_{ij}=0$. At entries $(i,j)\in\Omega$, the observations are independent draws 
as follows,
\begin{eqnarray}
\label{eq:Y-prob}
\ME{Y}{ij}&=&
\begin{cases}
+1& \text{with probability }\Phi(\ME{\theta}{ij}^*), \\
-1& \text{with probability }1-\Phi(\ME{\theta}{ij}^*),
\end{cases}   \amp \text{for} ~(i,j)\in \Omega
\end{eqnarray}
where $\ME{\theta}{ij}^*$ is the $ij$-th entry of $\M{\Theta}^*$, and $\Phi:\Real \rightarrow [0,1]$ is a known cumulative distribution function (CDF). 
Given 
the function $\Phi$ and the matrix $\M{Y}$, the goal of 1-bit matrix completion is to estimate the underlying matrix $\M{\Theta}^*$, 
see for example \cite{davenport20141}.

This problem is a variant of the standard matrix completion problem, in which one directly observes the entries $\theta_{ij}^*$ for $(i,j) \in \Omega$.
 In some applications, one can only observe a stochastic value $y_{ij}$ that is non-linearly related to the underlying value $\theta_{ij}^*$.  
 For example, in the Netflix challenge \citep{koren2009matrix}, the goal was to predict the missing entries of a rating matrix whose observed entries were integers ranging from one to five. In the simplest model for recommendation systems, user responses take on only one of two values:  ``like" or ``dislike", leading to a 1-bit matrix completion problem. 
 Additional applications involving completion of partially observed binary matrices include 
 quantum state tomography \citep{quantum} and link prediction in network analysis \citep{liben2003link}.


Given the partially observed binary matrix $\M{Y}$ and an assumed target rank $r$, perhaps the most natural approach 
to estimate $\M{\Theta}^*$ is by maximizing the likelihood function, or equivalently minimizing the negative log-likelihood $\ell(\M \Theta)$ under a rank-$r$ constraint,
\begin{eqnarray}
\label{eq:1-bit-opt}
\underset{\M{\Theta} \in \Real^{m \times n}}{\text{minimize}}\; \ell(\M{\Theta})\quad \text{subject to}\quad\text{rank}(\M{\Theta}) \, \leq \, r.
\end{eqnarray}
Specifically, under the generative model (\ref{eq:Y-prob}), 
$\ell(\M \Theta)$
is given by
\begin{eqnarray}
\label{eq:log-L}
    \ell(\M{\Theta}) &=& -\sum_{(i,j)\in \Omega} \bigg\{\ME{\delta}{ij} \log \Phi(\ME{\theta}{ij}) +(1-\ME{\delta}{ij})\log\left[1-\Phi(\ME{ \theta}{ij})\right]\bigg\},
\end{eqnarray}
where $\delta_{ij} = \frac{1}{2}(1 + \ME{Y}{ij})$  are rescalings of the observed entries.

\subsection{Prior Work}



Several authors derived algorithms as well as theoretical guarantees for the above 1-bit matrix completion problem. 
%
To ensure the 1-bit matrix completion problem is well-posed, assumptions are needed on  the underlying matrix $\M{\Theta}^*$ and on the subset $\Omega$ of observed entries. 
First, the entries of $\M\Theta^*$ cannot be too large in magnitude, so that the observed entries provide sufficient information on the underlying values of $\M\Theta^*$. Second, the rank of $\M\Theta^*$, or a proxy thereof, also cannot be too large. Finally, the matrix needs to be de-localized, and cannot be too spiky. 
Here, we adopt the notion 
of matrix spikiness from 
\cite{spikiness}. 
For a non-zero matrix $\M{\Theta} \in \Real^{m \times n}$, its spikiness ratio $s(\M{\Theta})$ is defined as  $s(\M{\Theta})  =  \sqrt{mn}\lVert \M{\Theta}\lVert_{\infty}/\lVert\M{\Theta} \lVert_{\text{F}}$
where $\lVert \M{\Theta} \rVert_\infty = \max_{ij} |\theta_{i,j}|$ and $\lVert \M{\Theta} \rVert_{\text{F}}$ is its Frobenius norm.
By definition, the spikiness ratio satisfies $1\leq s(\M{\Theta}) \leq \sqrt{mn}$. In general, a lower value of $s(\M{\Theta}^*)$ yields an easier 1-bit matrix completion problem.  

 
Additionally, 
 the set $\Omega$ of observed entries needs to be well spread-out. 
Several previous works assumed that $\Omega$ is distributed uniformly at random. Then if $|\Omega|$ is sufficiently large and the above assumptions on $\M{\Theta}^*$ hold, it is possible to accurately estimate $\M{\Theta}^*$, see for example \cite{bhaskar20151}, \cite{davenport20141}, and \cite{cai2013max}. 


%
%
Instead of the original problem \eqref{eq:1-bit-opt}, some works minimized 
(\ref{eq:log-L}) subject to a constraint on the rank and an additional constraint on $\lVert \M{\Theta}\lVert_{\infty}$ \citep{bhaskar20151, ni2016optimal}.
%
%
Other works employed a constraint on a convex relaxation of the rank in conjunction with the constraint on $\lVert \M{\Theta}\lVert_{\infty}$. For example, \cite{davenport20141} replaced the rank by the trace norm $\lVert\M{\Theta}\lVert_{*}$. They minimized the negative log-likelihood \eqref{eq:log-L} under two constraints $\lVert\M{\Theta}\lVert_{*} \leq  \alpha\beta\sqrt{mn}$ and $\lVert \M{\Theta}\lVert_{\infty} \leq \alpha$, where $\alpha$ and $\beta$ are two tuning parameters. Instead of the trace norm,
\citet{cai2013max} employed the max-norm \citep{linial2007complexity} as a convex relaxation of the rank. They minimized the objective \eqref{eq:log-L}
under the two constraints $\lVert\M{\Theta}\lVert_{\max}  \leq  \alpha\beta$ and $\lVert \M{\Theta}\lVert_{\infty}  \leq  \alpha$, where $\lVert \M{\Theta} \lVert_{\text{max}} = \min_{\M{\Theta}= \M U \M V \Tra} \{\lVert \M U \lVert_{2, \infty}, \lVert \M V \lVert_{2, \infty}\}$ and $\lVert \M U \lVert_{2, \infty}$ is the largest $l_2$-norm of the rows in $\M U$.

Most algorithms for these prior works, however, are computationally intensive and non-trivial to implement. \cite{bhaskar20151} employed a log-barrier penalty to approximate the infinity norm constraint in combination with a cross-validation procedure to determine how to scale the penalty.
The algorithm in \cite{davenport20141}, at each iteration, runs an inner ADMM algorithm to compute the projection onto the intersection of an infinity norm ball and nuclear norm ball. The algorithm in \cite{cai2013max}, at each iteration, requires projection onto a max-norm constraint which involves solving a nonconvex optimization problem iteratively in conjunction with a rank selection step.
Hence some of these methods do not scale well to large matrices and are prohibitively slow. As illustrated later on in Section \ref{sec: simulation}, publicly available implementations may take hours to complete medium-sized matrices.  


\subsection{Our Contributions}

The main contribution of this work is the derivation of a simple and computationally fast approach to minimize \eqref{eq:1-bit-opt}. 
Our iterative method, denoted Majorization-Minimization Gauss-Newton (\texttt{MMGN}), relies on a majorization-minimization strategy. This leads to solving a sequence of standard simpler matrix completion problems. 
We approximately solve these problems by a single Gauss-Newton step, each of which requires solving a least squares problem. Hence, our method is simple to implement, requiring only standard linear algebra tools. 
\texttt{MMGN} is typically at least an order of magnitude faster than existing 1-bit matrix completion methods while achieving comparable or even superior estimation accuracy. 
In addition, \texttt{MMGN} is faster than 
generic optimization methods that minimize the original objective \eqref{eq:1-bit-opt}, including
gradient descent, quasi-Newton methods, and manifold optimization algorithms. Beyond its computational advantages, as illustrated via simulations, \texttt{MMGN}'s performance degrades more slowly as the spikiness of the underlying matrix increases compared to alternative methods. 

\subsection{Notation and Organization}

We denote vectors by boldface lowercase letters and matrices by boldface capital letters, e.g.,  $\V a \in \Real^n$ and
$\M A \in \Real^{m \times n}$. We denote the entries of a vector $\V{a}$ and matrix $\M{A}$ by $\VE{a}{i}$ and $\ME{A}{ij}$, respectively. The rank of a matrix $\M A$ is denoted by $\text{rank}(\M A)$, 
its Moore–Penrose pseudo inverse by $\M A^\dagger$, and its column-major vectorization, i.e., 
the vector obtained by stacking the columns of $\M A$ one after the 
other, by $\text{vec}(\M A)$. We will 
use the following semi-norm of a matrix $\M{A}$, denoted by $\lVert \M{A}\rVert_{\text{F}(\Omega)} = \sqrt{\sum_{(i,j)\in\Omega} a_{ij}^2}$, where $\Omega$ is a subset of $\M{A}$'s indices. 
Similarly, for a vector $\V a$, we denote  $\lVert \V a \rVert_{\Omega}= \sqrt{\sum_{i \in \Omega} a_i^2}$, where $\Omega$ is a subset of $\V{a}$'s indices. For two matrices $\M A$ and $\M B$ of the same size, we denote the Hadamard product by $\M A \circ \M B$, namely $(\M A \circ \M B)_{ij} = \ME{A}{ij} \ME{B}{ij}$. 
Similarly, the element-wise quotient of $\M{A}$ and $\M{B}$ is $\M{A} \oslash \M{B}$, namely $(\M A \oslash \M B)_{ij} = \ME{A}{ij} / \ME{B}{ij}$. For a set $S$, we denote its cardinality by $|S|$. Finally, for any univariate function $\Phi : \Real \rightarrow \Real$,  we denote the matrix obtained by applying $\Phi$ entry-wise to $\M{A}$ by $\Phi(\M{A})$.

The rest of the paper is organized as follows. Section~\ref{sec: method} introduces the \texttt{MMGN} method. 
Section \ref{sec:related_problems} discusses several related problems and approaches similar to ours to solve them. 
Section~\ref{sec: simulation}  presents an empirical evaluation of \texttt{MMGN} with simulations and a real data example. Section~\ref{sec: discussion} concludes with a discussion.

\section{The \texttt{MMGN} Method}
\label{sec: method}

We first present an overview of our approach. 
Given a target rank $r$, our strategy for solving the nonconvex problem (\ref{eq:1-bit-opt}) is based on the majorization-minimization (MM) principle. 
Concretely, we replace the objective $\ell(\V{\Theta})$ with a simpler surrogate function, called a majorization. 
As we show below, minimizing our majorization leads to a standard low-rank matrix completion problem. Hence, \texttt{MMGN} solves a sequence of optimization problems of the form
\begin{eqnarray}
\label{eq:MM}
\underset{\M{\Theta} \in \Real^{m \times n}}{\text{minimize}}\; 
\left\lVert \M{\Theta} - \Mtilde{Y} \right\rVert_{\text{F}(\Omega)}^2 \quad\quad\text{subject to}\quad\quad \text{rank}(\M{\Theta}) \ampb \leq \ampb r,
\end{eqnarray}
where the matrix $\Mtilde{Y}$  depends on the observed matrix $\M{Y}$ and on the current estimate $\Mtilde{\Theta}$.

We solve (\ref{eq:MM}) by a factorization approach, expressing $\M{\Theta} = \M{U}\M{V}\Tra$ where $\M{U}\in \Real^{m \times r}$
and $\M{V}\in \Real^{n \times r}$.  Therefore, we rewrite (\ref{eq:MM}) as the following equivalent unconstrained problem
\begin{eqnarray}
\label{eq:problem_parameterization}
\underset{\M{U} \in \Real^{m \times r}, \M{V} \in \Real^{n \times r}}{\text{minimize}}\;
\left\lVert \M{U}\M{V}\Tra - \Mtilde{Y} \right\rVert_{\text{F}(\Omega)}^2.
\end{eqnarray}
Finally, we solve \Eqn{problem_parameterization} using a modified Gauss-Newton algorithm introduced by \cite{ZilberNadler2022}. 

\subsection{The Majorization-Minimization Principle}
\label{sec2.1-MM}

The MM principle \citep{deLeeuw1994,Heiser1995,lange2000optimization, HunterLange2004, lange2016mm} converts the minimization of a challenging function $\ell(\M{\Theta})$ into a sequence of simpler optimization problems. 
We approximate $\ell(\M{\Theta})$ by a surrogate function or majorization $g(\M{\Theta} \mid \Mtilde{\Theta})$ anchored at the current estimate $\Mtilde{\Theta}$. The majorization  $g(\M{\Theta} \mid \Mtilde{\Theta})$ needs to satisfy two conditions: (i) a tangency condition $g(\Mtilde{\Theta} \mid \Mtilde{\Theta}) = \ell(\Mtilde{\Theta})$ for all $\Mtilde{\Theta}$ and (ii) a domination condition $g(\M{\Theta} \mid \Mtilde{\Theta}) \geq \ell(\M{\Theta})$ for all $\M{\Theta}$. The associated MM algorithm is defined by the iterates
\begin{eqnarray}
\label{eq:MM-iterate}
    \M{\Theta}_{t+1} & = & \operatorname*{arg\,min}_{\M{\Theta}}  g(\M{\Theta} \mid \M{\Theta}_{t}), ~~~ t=0, 1, \dots.
\end{eqnarray}
The tangency and domination conditions imply that
\begin{eqnarray*}
    \ell(\M{\Theta}_{t+1}) ~~ \leq ~~ g(\M{\Theta}_{t+1} \mid \M{\Theta}_{t}) ~~  \leq ~~ g(\M{\Theta}_{t} \mid \M{\Theta}_{t}) ~~ = ~~ \ell(\M{\Theta}_{t}).
\end{eqnarray*}
In other words, the objective function values of the MM iterates decrease monotonically. Finding the global minimizer of $g(\M{\Theta} \mid \M{\Theta}_{t})$ is not necessary to ensure the descent property. Therefore, one can inexactly solve (\ref{eq:MM-iterate}) and still guarantee a monotonic decrease of the objective. A key component for the success of the MM principle is the construction of a surrogate function that is easy to minimize. 

Recall that we seek to minimize the negative log-likelihood of the 1-bit matrix completion problem  (\ref{eq:log-L}), which we restate here for convenience:
\begin{eqnarray*}
    \ell(\M{\Theta}) &=& -\sum_{(i,j)\in \Omega} \bigg\{\ME{\delta}{ij} \log \Phi(\ME{\theta}{ij}) +(1-\ME{\delta}{ij})\log\left[1-\Phi(\ME{ \theta}{ij})\right]\bigg\},
\end{eqnarray*}
where $\delta_{ij} = \frac{1}{2}(1 + \ME{Y}{ij}) \in \{0, 1\}$. 
To derive our majorization, we assume the CDF $\Phi(\theta)$ in (\ref{eq:log-L}) satisfies the following two conditions.
\begin{itemize}
\item[\textbf{A1.}]  The function $\log \Phi(\theta)$ is $L$-Lipschitz differentiable.
\item[\textbf{A2.}]  The density function $\phi(\theta) = \Phi'(\theta)$ is symmetric around zero. This implies
that $\Phi(\theta) = 1 - \Phi(-\theta)$.
\end{itemize}
The following proposition describes a quadratic majorization under the above assumptions.

\begin{proposition}
\label{prop:majorization-general}
Let $\Phi(\theta)$ be a CDF that satisfies assumptions A1 and A2. The following is a majorization of $\ell(\M{\Theta})$ at $\Mtilde{\Theta}$
\begin{eqnarray}
    g(\M{\Theta} \mid \Mtilde{\Theta}) &=& \frac{L}{2}\left \lVert \M{\Theta} - \Mtilde{Y}
    \right\rVert_{\text{F}(\Omega)}^2+c(\Mtilde{\Theta}), 
    \label{eq:general-quadratic-major}
\end{eqnarray}
where $c(\Mtilde{\Theta})$ depends on $\Mtilde{\Theta}$ but not on $\M{\Theta}$, and
\begin{eqnarray}
\label{eq:general-Y}
\Mtilde{Y} & = & \Mtilde{\Theta} + \frac{1}{L} \left(\M{Y}\circ \phi(\Mtilde{\Theta})\right) \oslash \Phi\left(\M{Y}\circ\Mtilde{\Theta}\right).
\end{eqnarray}
\end{proposition}

Two popular CDFs in 1-bit matrix completion are $\Phi(x) = 1/(1+e^{-x/\sigma})$ and $\Phi(x) = \frac{1}{\sqrt{2\pi\sigma^2}}\int_{-\infty}^x \exp\left(-\frac{w^2}{2\sigma^2} \right)dw$, which correspond to logistic and Gaussian random variables, respectively. 
Both satisfy assumptions A1 and A2. Hence, 
by 
\Prop{majorization-general} we obtain the following specific quadratic majorizations.

\begin{corollary}
\label{cor:majorization-logit}
    The logistic model with $\Phi(x)=1/(1+e^{-x/\sigma})$
    satisfies assumptions A1 and A2 with Lipschitz constant $L=\frac{1}{4\sigma^2}$. Hence, a majorization is given by Eq. \eqref{eq:general-quadratic-major}
    with 
\begin{eqnarray*}
\Mtilde{Y} & = & \Mtilde{\Theta}+4\sigma\M{Y} \circ  \Phi\left(-\M{Y} \circ \Mtilde{\Theta}\right).
\end{eqnarray*}
\end{corollary}

\begin{corollary}
\label{cor:majorization-probit}
    The probit model with $\Phi(x)=\frac{1}{\sqrt{2\pi\sigma^2}}\int_{-\infty}^x \exp\left(-\frac{w^2}{2\sigma^2} \right)dw$
    satisfies assumptions A1 and A2 with Lipschitz constant $L=\frac{1}{\sigma^2}$. Hence, a majorization is given by Eq. \eqref{eq:general-quadratic-major}
    with 
\begin{eqnarray*}
\Mtilde{Y} & = & \Mtilde{\Theta} + \sigma^2\M{Y}\circ \phi(\Mtilde{\Theta})\oslash \Phi(\M{Y}\circ \Mtilde{\Theta}).
\end{eqnarray*}
\end{corollary}


Proofs of \Prop{majorization-general}, \Cor{majorization-logit}, and \Cor{majorization-probit} are in the supplementary materials. 
Combined with the low-rank constraint, 
these corollaries indicate that both the logistic and probit models lead to MM updates that require solving a standard low-rank matrix completion problem.
\begin{eqnarray}
\label{eq:MM-update}
\underset{\M{\Theta}\in \Real^{m \times n}}{\text{minimize}}\;\left\lVert \M{\Theta} - \Mtilde{Y} \right\rVert_{\text{F}(\Omega)}^2 \quad\quad\text{subject to}\quad\quad \text{rank}(\M{\Theta}) \amp \leq \amp r.
\end{eqnarray}
We denote the solution to the problem (\ref{eq:MM-update}) by $\M{\Theta}_{\text{MM}}$.
We next review how we approximately solve the non-convex problem (\ref{eq:MM-update}) by a Gauss-Newton strategy.

\subsection{Inexact Majorization-Minimization via Gauss-Newton}
\label{sec2.2-GN}

Let $g(\M{\Theta} \mid \Mtilde{\Theta})$ be a quadratic majorization of the form  (\ref{eq:general-quadratic-major}) with $\Mtilde{Y}$ defined in (\ref{eq:general-Y}), where 
$\Mtilde{\Theta}$ is the current estimate of $\M{\Theta}$. 
It is useful to express $g(\M{\Theta} \mid \Mtilde{\Theta})$ in terms of the difference $\M{\Theta} - \Mtilde{\Theta}$.  The MM-update in (\ref{eq:MM-update}) can then be written as 
\begin{eqnarray}
\label{eq:MM-update2}
\M{\Theta}_{\text{MM}} & = & 
\underset{\textrm{rank}(\M{\Theta})\leq r}{\arg\min}\; g(\M{\Theta}|\Mtilde{\Theta}) \amp = \amp \underset{\textrm{rank}(\M{\Theta})\leq r}{\arg\min}\; \left \lVert\M{\Theta}-\Mtilde{\Theta}-\M{X}\right \rVert_{\text{F}(\Omega)}^2,
\label{eq:LS-M}
\end{eqnarray}
where
\begin{eqnarray}
\label{eq:M_X}
\M{X} & = & \Mtilde{Y} - \Mtilde{\Theta} \amp = \amp \frac{1}{L} \M{Y}\circ \phi(\Mtilde{\Theta}) \oslash \Phi\left(\M{Y}\circ\Mtilde{\Theta}\right).
\end{eqnarray}
We solve (\ref{eq:MM-update2}) by the following factorization approach. We express $\Mtilde{\Theta}$ and $\M{\Theta}$ as the products of two rank-$r$ factor matrices with 
\begin{eqnarray*}
\Mtilde{\Theta} & = & \tilde{\M U} \tilde{\M V}\Tra \quad\quad \text{and} \quad\quad \M{\Theta} \amp = \amp \left(\Mtilde{U} +\Delta\M{U}\right)\left(\tilde{\M{V}}+\Delta\M{V}\right)\Tra.
\end{eqnarray*}
The variables $\Mtilde{U} \in \Real^{m \times r}$ and $\Mtilde{V} \in \Real^{n \times r}$ are the factor matrices corresponding to the current estimate $\Mtilde{\Theta}$, whereas $\Delta\M{U} \in \Real^{m \times r}$ and $\Delta\M{V} \in \Real^{n \times r}$ are their updates. 
Therefore, 
problem (\ref{eq:LS-M}) can be  written in terms of the new variables ($\Delta \M U$, $\Delta \M V$) 
as
\begin{eqnarray}
\label{eq:nonlinear_least_squares}
     \underset{\Delta \M{U},\Delta\M{V}}{\text{minimize}}\;\left\lVert \M{X}-\tilde{\M U}\Delta\M{V}\Tra - \Delta\M{U}\tilde{\M{V}}\Tra -
     \Delta\M{U} \Delta\M{V}\Tra
     \right\rVert_{\text{F}(\Omega)}^2.
\end{eqnarray}

The optimization problem in (\ref{eq:nonlinear_least_squares}) is a nonlinear least squares problem. 
Motivated by \citet{ZilberNadler2022}, 
 we neglect the second order term $\Delta\M{U} \Delta\M{V}\Tra$. We thus  compute the solution ($\Delta \M U^*$, $\Delta \M V^*$) to
\begin{eqnarray}
\label{eq:LS-UV}
     \underset{\Delta \M{U},\Delta\M{V}}{\text{minimize}}\;\left\lVert\M{X}-\tilde{\M{U}}\Delta\M{V}\Tra - \Delta\M{U}\tilde{\M{V}}\Tra
     \right\rVert_{\text{F}(\Omega)}^2.
\end{eqnarray}
Approximating the nonlinear least squares problem with the linear least squares problem in (\ref{eq:LS-UV}) corresponds to taking a single Gauss-Newton iteration.
This approach produces the following inexact solution of (\ref{eq:LS-M}), denoted $\hat{\M{\Theta}}_{\text{MM}} $, 
\begin{eqnarray}
\label{eq:MM-update-matrix}
   \hat{\M{\Theta}}_{\text{MM}} & = & \;\; \left(\tilde{\M U}+ \Delta \M U^*\right)\left(\tilde{\M V} +  \Delta \M V^*\right)\Tra.
\end{eqnarray}

We address three important details about our approach. First, the linear least squares problem (\ref{eq:LS-UV}) has infinitely many solutions. For example, suppose that $(\Delta \M{U}^*, \Delta\M{V}^*)$ is a solution to (\ref{eq:LS-UV}), then $(\Delta \M{U}^* + \Mtilde{U}\M{R}, \Delta\M{V}^* - \Mtilde{V}\M{R}\Tra)$ is also a solution for any $\M{R} \in \Real^{r \times r}$. Here we
adopt the strategy used by \cite{bauch2021rank} and \cite{ZilberNadler2022} and select the least $l_2$-norm solution, namely $(\Delta \M{U}^*, \Delta \M{V}^*)$ with smallest value of $\lVert \Delta \M{U}^* \rVert_{\text{F}}^2 + \lVert \Delta \M{V}^* \rVert_{\text{F}}^2$ among all $(\Delta \M{U}^*, \Delta \M{V}^*)$ that solve (\ref{eq:LS-UV}). This solution can be computed efficiently using the LSQR algorithm~\citep{PaigeSaunders1982} or the conjugate gradient method \citep{hestenes1952methods, kammerer1972convergence} applied to the normal equations.
 
The second detail is that the update in \Eqn{MM-update-matrix} may not necessarily lead to a decrease in the objective function. 
The Gauss-Newton method is an instance of the steepest descent algorithm, and consequently the solution  $(\Delta \M{U}^*, \Delta \M{V}^*)$ corresponds to a descent direction. In fact, we prove in the supplementary materials that any solution to \Eqn{LS-UV} is a descent direction. The update in \Eqn{MM-update-matrix}, however, corresponds to a full Gauss-Newton step which may not necessarily decrease the value of the original objective function in (\ref{eq:LS-M}). To overcome this potential problem, if  $\hat{\M{\Theta}}_{\text{MM}}$ in \Eqn{MM-update-matrix} does not decrease the original objective, we apply the Armijo backtracking line search to select a suitable stepsize \citep[Chapter  3]{NoceWrig06}.  The Armijo backtracking line search requires computing the objective value, which costs $\mathcal{O}(|\Omega|r)$ flops. 
In our experience, however, the backtracking line search is seldom triggered in practice. Details on this and a discussion on the convergence of \texttt{MMGN} are in the supplementary materials.  


The third detail is that there is no need to globally minimize the majorization (\ref{eq:LS-M}). Motivated by computational considerations, we inexactly solve the MM optimization problem by taking a {\em single} Gauss-Newton step. This differs from the approach of 
\citet{ZilberNadler2022}, which solves \eqref{eq:LS-UV} using multiple iteration steps and is thus more computationally intensive. Inexact minimization is a standard approach in cases where exact minimization requires an iterative solver. See, for example, the split-feasibility algorithm in \cite{Xu2018}, which also employed a single Gauss-Newton step. Under suitable regularity conditions, little is lost by a single-step MM-gradient approach. Not only does it avoid potentially expensive inner iterations within outer MM iterations, but it often exhibits the same local convergence rate as exact minimization \citep{HunterLange2004}. 

\begin{algorithm}[th]
\caption{\texttt{MMGN}}
\label{alg:MMGN}
\begin{algorithmic}[1]
\Require $\Omega$, $\M{Y}$, target rank $r$, tolerance $\texttt{tol}$
\State Initialize $\M{U}_0$, $\M{V}_0$, set $t=0$,  $\mathrm{rel} = \texttt{INF}$
\While{$\mathrm{rel} > \texttt{tol}$} 
\State Construct $\M{X}_{t}$ in \eqref{eq:M_X} as \Comment{Majorization}
\begin{eqnarray*}
\M{X}_{t} & = & \frac{1}{L} \M{Y}\circ \phi\left(\M{U}_{t}\M{V}_{t}\Tra\right) \oslash \Phi\left(\M{Y}\circ \left(\M{U}_{t}\M{V}_{t}\Tra\right)\right).
\end{eqnarray*}
\State Compute the least $l_2$-norm solution $(\Delta\M{U}^*, \Delta\M{V}^*)$ to \Comment{Single MM-gradient step}
\begin{eqnarray*}
    \underset{\Delta \M{U},\Delta\M{V}}{\text{minimize}}\;\left\lVert \M{X}_{t}-\M{U}_{t}\Delta\M{V}\Tra - \Delta\M{U}\M{V}_{t}\Tra
     \right\rVert_{\text{F}(\Omega)}^2.
\end{eqnarray*} 
\State Update factor matrices
\begin{eqnarray*}
\begin{bmatrix}
\M{U}_{t+1} \\
\M{V}_{t+1} \\
\end{bmatrix}
& = & 
\begin{bmatrix}
\M{U}_{t} \\
\M{V}_{t}
\end{bmatrix}
+ \alpha_{t} 
\begin{bmatrix}
\Delta\M{U}^*\\
\Delta\M{V}^*
\end{bmatrix},
\end{eqnarray*}
where $\alpha_{t}$ is the stepsize chosen by  Armijo backtracking line search.
\State Compute the relative change in the log-likelihood
\begin{eqnarray*}
\mathrm{rel} & = & \frac{|\ell(\M{U}_{t+1}\M{V}_{t+1}\Tra)- \ell(\M{U}_{t}\M{V}_{t}\Tra) |}{| \ell(\M{U}_{t}\M{V}_{t} \Tra)|}.
\end{eqnarray*}
\State $t=t+1$.
\EndWhile \\
\Return $\hat{\M{\Theta}} = \M{U}_{t}\M{V}_{t}\Tra$.
\end{algorithmic}
\end{algorithm}

We present the complete \texttt{MMGN} algorithm in \Alg{MMGN}. \texttt{MMGN} requires $\mathcal{O}(|\Omega|r)$ operations per iteration, which is comparable to the gradient-based methods. Details on these per-iteration costs can be found in Section C of the supplement. Empirically, however, \texttt{MMGN} consistently outperforms generic gradient-based methods in terms of runtime as shown in Section \ref{sec: simulation}. We close this section with a brief discussion to understand why this might be expected.

For simplicity, we work with a bivariate instance of the objective function for the factorization approach. Let $\ell: \Real \mapsto \Real$ be twice differentiable with an $L$-Lipschitz first derivative. Define $f: \Real \times \Real \mapsto \Real$ as $f(u, v) = \ell(uv)$. Let $u = \tilde{u} + \Delta u$ and $v = \tilde{v} + \Delta v$ and 
consider the second order Taylor expansion of $f$ around $(\tilde{u}, \tilde{v})$.
\begin{eqnarray}
\label{taylor_expansion}
f(u, v) & = & 
\ell(\tilde{u}\tilde{v}) + \ell'(\tilde{u}\tilde{v})[\Delta u \tilde{v} + \tilde{u}\Delta v ]+ 
\frac{\ell''(\tilde{u}\tilde{v})}{2}[\Delta u \tilde{v} + \tilde{u} \Delta v]^2 + r(\Delta u, \Delta v),
\end{eqnarray}
where $r(\Delta u, \Delta v)$ is $o( (\Delta u + \Delta v)^3)$.

\texttt{MMGN} employs the following majorization of $f(u,v)$ at $(\tilde{u}, \tilde{v})$
\begin{eqnarray}
\label{mmgn_quadratic_majorization}
g(u,v \mid \tilde{u}, \tilde{v}) & = & \ell(\tilde{u}\tilde{v}) + \ell'(\tilde{u}\tilde{v})[\Delta u \tilde{v} + \tilde{u}\Delta v ]+ 
\frac{L}{2}[\Delta u \tilde{v} + \tilde{u} \Delta v]^2.
\end{eqnarray}
There are two differences between \eqref{taylor_expansion} and \eqref{mmgn_quadratic_majorization}. First, \texttt{MMGN} uses the uniform bound $L$ in lieu of computing the second derivative $\ell''(\tilde{u}\tilde{v})$. This substitution greatly simplifies computations in the general case when $\ell$ is a function of matrices $\M{U}\in \Real^{m \times r}$ and $\M{V} \in \Real^{n \times r}$. Second, \texttt{MMGN} drops the third order error term $r(\Delta u, \Delta v)$. Recall that the minimizer to \eqref{mmgn_quadratic_majorization} is not unique and that \texttt{MMGN} chooses the least 2-norm minimizer. This has the beneficial side-effect of finding a descent direction along which the error term $r(\Delta u, \Delta v)$ is small. In other words, \texttt{MMGN} picks a descent direction, among many possible descent directions, along which the surrogate function has a small approximation error. This small error is likely the reason the full Gauss-Newton step is almost always taken by \texttt{MMGN}.

By contrast, the gradient descent surrogate is
\begin{eqnarray}
\label{gd_quadratic_majorization}
h(u,v \mid \tilde{u}, \tilde{v}) & = & \ell(\tilde{u}\tilde{v}) + \ell'(\tilde{u}\tilde{v})[\Delta u \tilde{v} + \tilde{u}\Delta v ]+ 
\frac{1}{2\alpha}[\Delta u^2 + \Delta v^2],
\end{eqnarray}
where $\alpha$ is the step-size parameter.

There are again two differences to note. First, \eqref{gd_quadratic_majorization} uses the isotropic quadratic function
\begin{eqnarray*}
\frac{1}{2\alpha}(\Delta u^2 + \Delta v^2)
\end{eqnarray*}
to approximate the anisotropic quadratic function
\begin{eqnarray*}
\frac{\ell''(\tilde{u}\tilde{v})}{2}(\Delta u \tilde{v} + \tilde{u}\Delta v)^2
\end{eqnarray*}
in \eqref{taylor_expansion}. Moreover, the quadratic term in \eqref{gd_quadratic_majorization} does not even depend on the current iterate $(\tilde{u}, \tilde{v})$. By contrast, \texttt{MMGN} uses a similar anisotropic quadratic function in \eqref{mmgn_quadratic_majorization} as in \eqref{taylor_expansion}. The quadratic dependency in \eqref{mmgn_quadratic_majorization} on $(\tilde{u}, \tilde{v})$ and $(\Delta u, \Delta v)$ is the same up to a scalar multiple as in \eqref{taylor_expansion}.

Second, the update direction $(\Delta u, \Delta v)$ chosen by gradient descent can be large when far from a solution. Thus, $r(\Delta u, \Delta v)$ could be large and so the surrogate approximation in \eqref{gd_quadratic_majorization} may be poor along the gradient direction. This is a well known issue with gradient descent that typically results in gradient descent taking very small steps when the objective function has steep and narrow valleys.

Now recall that the per-iteration cost of \texttt{MMGN} is $O(\lvert \Omega \rvert r)$. The cost of computing the gradient of $\ell(\M{U}\M{V}\Tra)$ is also $O(\lvert \Omega \rvert r)$. The above simple analysis suggests that \texttt{MMGN} computes updates from surrogate models that capture useful second order information with a per-iteration cost that is roughly the same cost as a gradient evaluation. In other words, we expect \texttt{MMGN} to behave as a quasi-Newton method that has a per-iteration cost similar to a first order method.

\section{Related Problems}
\label{sec:related_problems}

The 1-bit matrix completion problem as well as our approach to solving it are related to several other problems, as we outline below. 

\paragraph{Logistic PCA.}  

In logistic PCA, the entries of a binary data matrix $\M{Y} \in \{-1, 1\}^{m \times n}$ are modeled as independent draws $y_{ij} \sim \text{Bernoulli}(\Phi(\theta_{ij}^*))$ 
where $\M{\Theta}^* \in \Real^{m \times n}$ is rank-$r$ \citep{collins2001generalization, schein2003generalized}.  Logistic PCA shares the same modeling assumptions with 1-bit matrix completion but differs in two ways. 
First, as its name suggests, in 1-bit matrix completion,  only part of the data matrix is observed, while in logistic PCA, all matrix entries are observed.
Second, logistic PCA and 1-bit matrix completion have somewhat different goals. The main purpose of the former is dimension reduction, while the latter is prediction of the most likely values at the unobserved entries.

Maximum likelihood algorithms \citep{collins2001generalization, schein2003generalized, de2006principal} have been developed for logistic PCA, which share the same objective function \eqref{eq:log-L}. These approaches parameterize $\M \Theta$ with an explicit low-rank factorization, i.e., $\M \Theta = \M W \M H\Tra$. In particular, \cite{de2006principal} developed an MM algorithm for logistic PCA, which employed the same majorization of the objective \eqref{eq:log-L} as in our work. 
At each MM iteration, the corresponding rank-constrained least squares problem was solved using a weighted singular value decomposition (SVD). 
In principle, his approach may be adapted to the 1-bit matrix completion problem. 
However, as the simulations in the 
supplementary materials demonstrate, when the number of observed entries is small, our method 
is substantially faster. 

\paragraph{Binary matrix factorization.}

The goal of binary matrix factorization (BMF) is to recover a latent low-rank matrix of probabilities from a binary matrix that is typically fully observed \citep{bingham2009aspect,kaban2008factorisation, lumbreras2020bayesian, magron2022majorization}. In the BMF problem, the entries are independent draws from a Bernoulli distribution, i.e., $ y_{ij} \sim \text{Bernoulli}([\M W\M H\Tra]_{ij})$ where $\M W \in \Real^{m \times r}$  and $\M H \in \Real^{n \times r}$. That is, BMF assumes an identity link function $\Phi(\theta) = \theta$. This is in contrast to a distribution function $\Phi$ for the link in 1-bit matrix completion. In BMF, to guarantee that the entries of  $\M{W}\M{H}\Tra$ are probabilities, additional constraints are imposed on the factor matrices, e.g., the rows of $\M{W}$ are assumed to be stochastic and the entries of $\M{H}$ binary.

One of the main uses of BMF is dimensionality reduction for exploratory analysis. The identity link is employed because it produces factor matrices with better interpretability. The additional constraints accompanying the identity link, however, lead to a more complicated computational task. Despite having a somewhat different goal, BMF methods could be used to solve the 1-bit matrix completion problem. Simulations in the supplementary materials show that \texttt{MMGN} is typically more accurate and faster than the state-of-the-art BMF method \texttt{NBMF-MM} \citep{magron2022majorization}.

\paragraph{Link prediction in network analysis.}

In an unweighted undirected network, links between nodes are represented by a binary adjacency matrix. This binary adjacency matrix may be partially observed. A fundamental task in network analysis is to predict whether a link is present at a corresponding unobserved entry in the adjacency matrix  \citep{liben2003link}. A variety of link prediction methods have been developed, among which the generative model-based methods \citep{miller2009nonparametric, zhou2015infinite, zhou2015nonparametric} share similarities with 1-bit matrix completion. To model the binary link $y_{ij}$ between nodes $i$ and $j$ in a network with $n$ nodes, they assume $y_{ij} \sim \text{Bernoulli}(\Phi([\M Z \M W \M Z\Tra]_{ij}))$ where $\M Z \in \Real^{n \times k}$, $\M W \in \Real^{k \times k}$, and $\Phi$ is a link function. Assumptions on $\M Z$, $\M W$, and $\Phi$ vary across different models. \cite{miller2009nonparametric} assumed $\M Z$ to be a binary latent feature matrix. \cite{zhou2015infinite, zhou2015nonparametric} assumed a Bernoulli-Poisson link $\Phi(\theta) = 1-e^{-\theta}$, which implies that if the binary observation $y_{ij}=0$, then the latent mean parameter $[\M X \M W \M Z\Tra]_{ij} =0$. Thus, a key difference between link prediction and 1-bit matrix completion is that for some observed matrix entries, there is no uncertainty in the underlying latent mean parameter matrix.

\section{Numerical Studies}
\label{sec: simulation}

We compared \texttt{MMGN}\footnote{We implemented \texttt{MMGN} in both Matlab and R. All codes and simulation scripts are available at  \href{https://github.com/Xiaoqian-Liu/MMGN}{https://github.com/Xiaoqian-Liu/MMGN}. }
to two 1-bit matrix completion methods, \texttt{TraceNorm} \citep{davenport20141} and \texttt{MaxNorm}  \citep{cai2013max}, on 
simulated data as well as a real data example. We used the computationally more efficient version of  \texttt{TraceNorm} that omits the infinity norm constraint $\lVert \M{\Theta} \lVert_{\infty} \leq \alpha$. Although \cite{davenport20141}  
 required the constraint to establish their error bound, they observed that omitting it did not significantly affect the estimation performance and simplified the optimization problem. The methods in \cite{bhaskar20151} and \cite{ni2016optimal} have no publicly available code and consequently were not included in our comparisons. We used  
 MATLAB implementations of \texttt{TraceNorm} and \texttt{MaxNorm} provided by their authors. 
 
In addition, the original problem \eqref{eq:1-bit-opt} can be solved by generic optimization approaches. Thus, we also compared \texttt{MMGN} to three general schemes: gradient descent (\texttt{GD}) with backtracking line search as in \texttt{MMGN}, limited-memory BFGS (\texttt{LBFGS}), and \texttt{Manopt} \citep{manopt} for optimization on the Riemannian manifold of fixed-rank matrices. Details on the implementations appear in the supplementary materials. We also applied \texttt{GD} and \texttt{LBGS} to ridge regularized version of 1-bit matrix completion and report results on these variations in the supplementary materials.

\subsection{Simulation Settings}
\label{sec: 3.1}

\noindent\textbf{Data generation.} In our simulations, we considered two types of  underlying low-rank matrices: non-spiky and spiky. We generated matrices of each type as follows.
\begin{itemize}
    \item \textbf{Non-spiky.} To generate a non-spiky matrix $\M{\Theta}^*\in \Real^{m\times n}$ of rank $r^*$, we constructed $\M{U}^*\in \Real^{m\times r^*}$, $\M{V}^* \in \Real^{n\times r^*}$ with independent and identically distributed (i.i.d.\@) entries from a uniform distribution on $[-0.5, 0.5]$. We set $\M{\Theta}^* = \M{U}^*\M{V}^{\sf *T}$ and scaled it so that $\|\M{\Theta}^*\|_{\infty} = 1$. This data generation procedure was the same as in \cite{davenport20141} and \cite{cai2013max}.
    \item \textbf{Spiky.} Here, we constructed $\M{U}^* \in \Real^{m \times r^*}$, $\M{V}^* \in \Real^{n \times r^*}$  with i.i.d.\@ entries from the $t$-distribution with $\nu$ degrees of freedom. We set $\M{\Theta}^* = \M{U}^*\M{V}^{\sf *T}$, without rescaling. 
    A smaller value of $\nu$ yielded a heavier-tailed distribution, resulting in a spikier matrix.
\end{itemize}
For each simulation, we first generated an underlying low-rank matrix $\M{\Theta}^*$. Then we randomly selected the set of indices $\Omega$ from a uniform distribution with a user-defined observation fraction $\rho$, namely $\rho = \frac{\lvert \Omega \rvert}{mn}$. Finally, we produced the binary matrix $\M Y$ according to the model (\ref{eq:Y-prob}).  

\noindent{\textbf{Rank estimation.}  In practice, we need to estimate the rank from the data.
 In all simulations, we used the following validation approach to estimate the rank. Given a candidate set of $K$ ranks $\{r_1, \cdots, r_K\}$, we randomly split the observations into a training set  (80\%)  and a validation set (20\%). For each candidate rank $r_k$, we computed an estimate $\hat{\M \Theta}_k$ using the training set and then calculated the likelihood of $\hat{\M \Theta}_k$ on the validation set. We selected the candidate rank with the highest likelihood among all $K$ candidates as our estimated rank $r$. Even when given a rank $r$, \texttt{TraceNorm} and \texttt{MaxNorm} could output a matrix with a higher rank. Consequently, we computed a truncated rank-$r$ SVD on their outputs to return a rank-$r$ matrix.

\noindent\textbf{Performance evaluation.} We evaluated the methods with the following three metrics.
\begin{itemize}
    \item[1)] The relative error $\lVert \hat{\M{\Theta}}-\M{\Theta}^*\rVert_{\text{F}}^2/\lVert\M{\Theta}^*\rVert_{\text{F}}^2$.
    
    \item[2)] The Hellinger distance between the distribution $\Phi(\hat{\M \Theta})$ and the true distribution $\Phi(\M \Theta^*)$. Following \cite{davenport20141} and \cite{ cai2013max}, we define the Hellinger distance between two matrices $\M P, \M Q \in [0, 1]^{m \times n}$ as
    \begin{eqnarray*}
        d_{H}^2 (\M P, \M Q) & = & \frac{1}{mn} \sum_{i, j}d_H^2(\M P_{ij}, \M Q_{ij}),
    \end{eqnarray*}
    where $d_H^2(p, q) = (\sqrt{p} - \sqrt{q})^2  + (\sqrt{1-p} - \sqrt{1-q})^2 $ for $p, q \in [0, 1]$. The Hellinger distance $d_H^2(p, q)$ is a standard metric between two probability distributions. It is nonnegative and equals zero when the two distributions are identical. 
    It provides a different perspective on estimation accuracy compared to relative error as we will see later at the end of Section~\ref{simulation-spiky}.

    \item[3)] The runtime in seconds. 
    
\end{itemize}
Some comparison methods in our simulation studies had fairly long run times. Consequently, we ran only 20 replicates for each simulation setting. Additionally, performance measures for some of the comparison methods exhibited large variability. Consequently, to better visualize trends, the median as well as the 25th and 75th quantiles of metrics are shown using box plots.

\subsection{Non-spiky Matrices}
\label{simulation-nonspiky}

We first compared the performance of different methods when the underlying matrix is non-spiky. 
We considered the probit noise model.  We varied problem parameters, including the noise level $\sigma$, the observation fraction $\rho$, the matrix dimension $n$, and the true rank $r$. 

\subsubsection{Varying the noise level}
\label{sec: exp-noise}
As discussed in \cite{davenport20141}, the noise level $\sigma$ is a crucial parameter in 1-bit matrix completion. As $\sigma$ tends to zero, the problem becomes ill-posed, as two different positive values for $\theta_{ij}^*$ yield the same value $y_{ij}=1$.
The first simulation explores the sensitivity of different methods to the noise level $\sigma$. We varied $\sigma$ from $10^{-1.25}$ to $10^{0.25}$ over equally spaced values on a log-linear grid. We set the matrix dimension $m= n = 1000$ and rank $r^*=1$. The generated underlying matrices are non-spiky with an average spikiness ratio of $3.02$ and a standard deviation of $0.07$. For each method, we estimated its rank $r$ from $\{1,2,3,4,5\}$ by the validation procedure described in 
Section \ref{sec: 3.1}.

\Fig{nonspiky-noise} shows the box plots of relative errors, Hellinger distances, and runtimes (in seconds) of all methods with an observation fraction $\rho = 0.3$. 
The left panel shows that when the noise level is either too low or too high, all methods produced poor relative errors. This corroborates the results in \cite{davenport20141} and \cite{cai2013max}. \texttt{MaxNorm} performed slightly better than the other methods when the noise level was extremely low or high. \texttt{TraceNorm} suffered from the highest relative error over a wide range of noise levels. \texttt{MMGN} and the three generic schemes for solving problem \eqref{eq:1-bit-opt} achieved comparable errors when the noise level $\sigma \geq 10^{-0.75}$. The middle panel shows that the Hellinger distance of \texttt{TraceNorm} was significantly larger than those of the other methods, except for $\sigma=10^{-1.25}$ where the Hellinger distance of \texttt{LBFGS} was even larger. As seen in the left and middle panels of \Fig{nonspiky-noise}, the relative errors and Hellinger distances disagreed in their assessments of the methods. The two metrics capture different errors. We will discuss these differences later. The right panel shows that runtimes of all methods decreased as the noise level grew, and \texttt{MMGN} ran the fastest among all compared methods. \texttt{MaxNorm} was more than $20$ times slower than all the other methods over a wide range of noise levels. In light of this, in all subsequent simulations we supplied \texttt{MaxNorm} with the true rank $r^*$ instead of estimating it using the validation procedure.


\begin{figure}
    \centering \includegraphics[width=\textwidth]{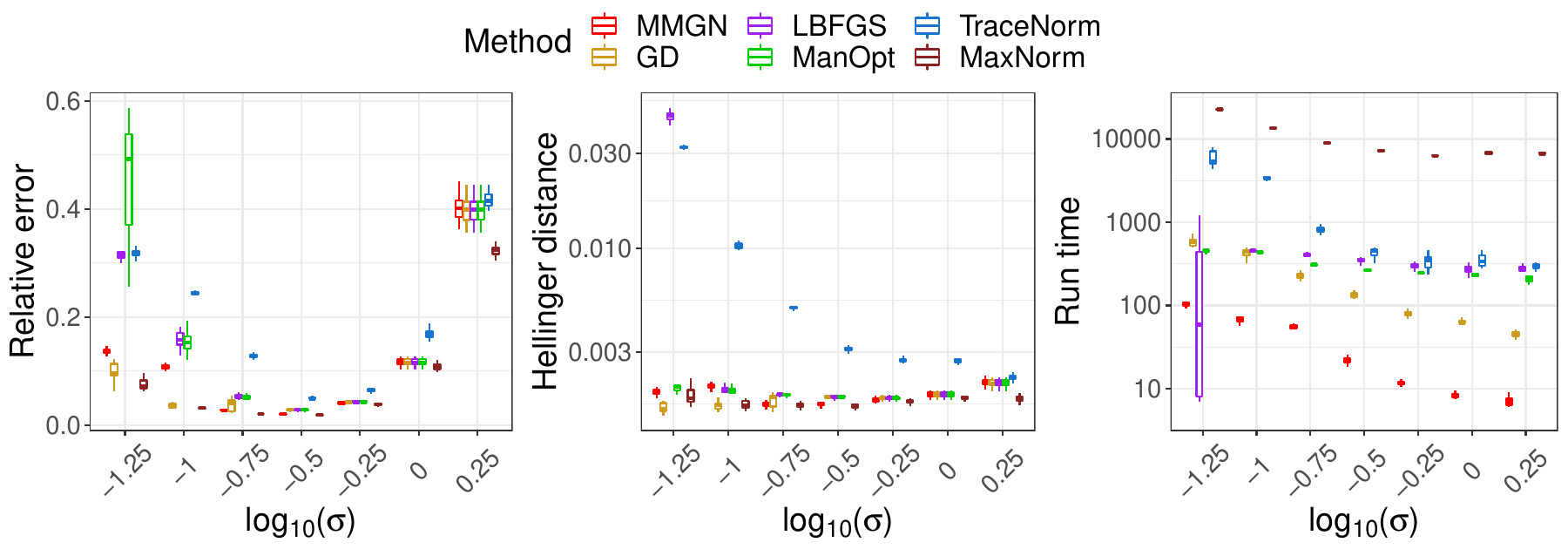}
    \caption{Probit model: Relative error, Hellinger distance, and runtime versus noise level $\sigma$ for a non-spiky underlying matrix of size $m \times n = 1000 \times 1000$ and rank $r^* = 1$ at observation fraction $\rho = 0.3$.
    }
    \label{fig:nonspiky-noise}
\end{figure}

\subsubsection{Varying the observation fraction}

In the second simulation, we investigated the performance of different methods as 
a function of the observation fraction $\rho \in \{0.2, 0.3, \ldots, 1\}$. The underlying matrices were the same as in the simulations in Section \ref{sec: exp-noise}.
We considered the probit model with a noise level $\sigma =1$. We estimated the rank $r$ from $\{1,2,3,4,5\}$ by the validation procedure described in Section \ref{sec: 3.1} for all methods except for \texttt{MaxNorm} which was provided with the true rank.


\begin{figure}
    \centering \includegraphics[width=\textwidth]{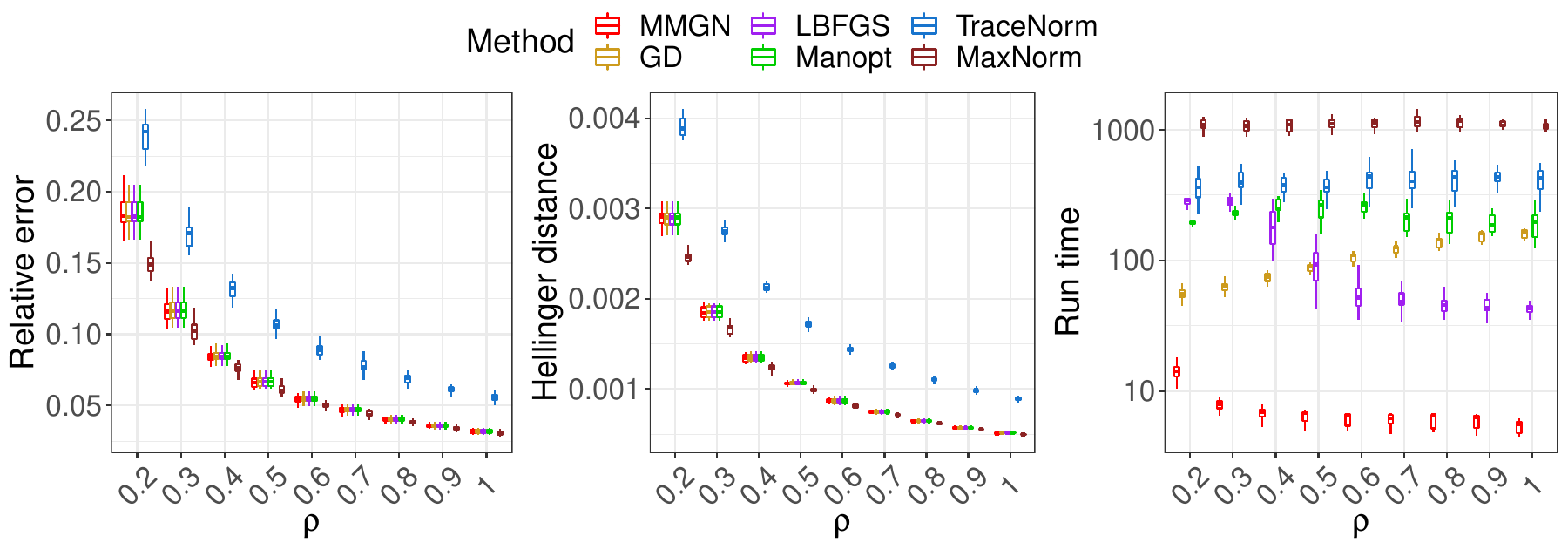}
    \caption{Probit model ($\sigma = 1$): Relative error, Hellinger distance, and runtime versus observation fraction $\rho$ for a non-spiky underlying matrix of size $m \times n = 1000 \times 1000$ and rank $r^* = 1$. 
    }
    \label{fig:nonspiky-probit-rho-r1}
\end{figure}

The results of this simulation are shown in \Fig{nonspiky-probit-rho-r1}. As expected, the relative errors as well as the Hellinger distances of all methods decreased as the observation fraction $\rho$ increased. Overall, \texttt{MMGN} behaved comparably with \texttt{MaxNorm} and consistently outperformed \texttt{TraceNorm}. Compared to the three generic schemes, \texttt{GD}, \texttt{LBFGS}, and \texttt{Manopt}, \texttt{MMGN} achieved similar estimation performance while running more than five times faster as seen in 
the right panel of \Fig{nonspiky-probit-rho-r1}. Moreover, \texttt{MMGN} required substantially less time to achieve comparable estimation accuracy with \texttt{MaxNorm} even when \texttt{MaxNorm} enjoyed the advantage of employing the true rank $r^*$. 

\subsubsection{Varying the matrix dimension}

Next, we compared the performance of different methods under different matrix dimensions. We considered  underlying square matrices of  dimension $n \in \{1000, 1500, 2000, 2500, 3000\}$, all with rank  $r^*=5$.
The average spikiness ratios of the generated matrices were $\{ 4.61, 4.81,$
$ 5.04, 5.09, 5.19, 5.22\}$ for the corresponding values of $n$. 
We considered the probit model with $\sigma =0.18$ and fixed the observation fraction $\rho=0.8$. 
For all methods other than \texttt{MaxNorm} which was given the true rank $r^*$, we employed the validation approach to estimate the rank $r$ from a candidate set $\{3, 4, 5, 6, 7\}$. The results are shown in \Fig{nonspiky-probit-dim-r1}.

Overall, all methods exhibited comparable accuracies that improved as the matrix dimension $n$ grew, with \texttt{TraceNorm}'s accuracy being slightly worse than the others. The right panel of \Fig{nonspiky-probit-dim-r1} shows \texttt{MMGN}'s computational advantage over all the other methods. \texttt{MMGN} was up to hundreds of times faster than 
\texttt{TraceNorm} and \texttt{MaxNorm}. In addition, it typically ran two to ten times faster than the three generic schemes for solving the optimization problem \eqref{eq:1-bit-opt}.


\begin{figure}
    \centering \includegraphics[width=\textwidth]{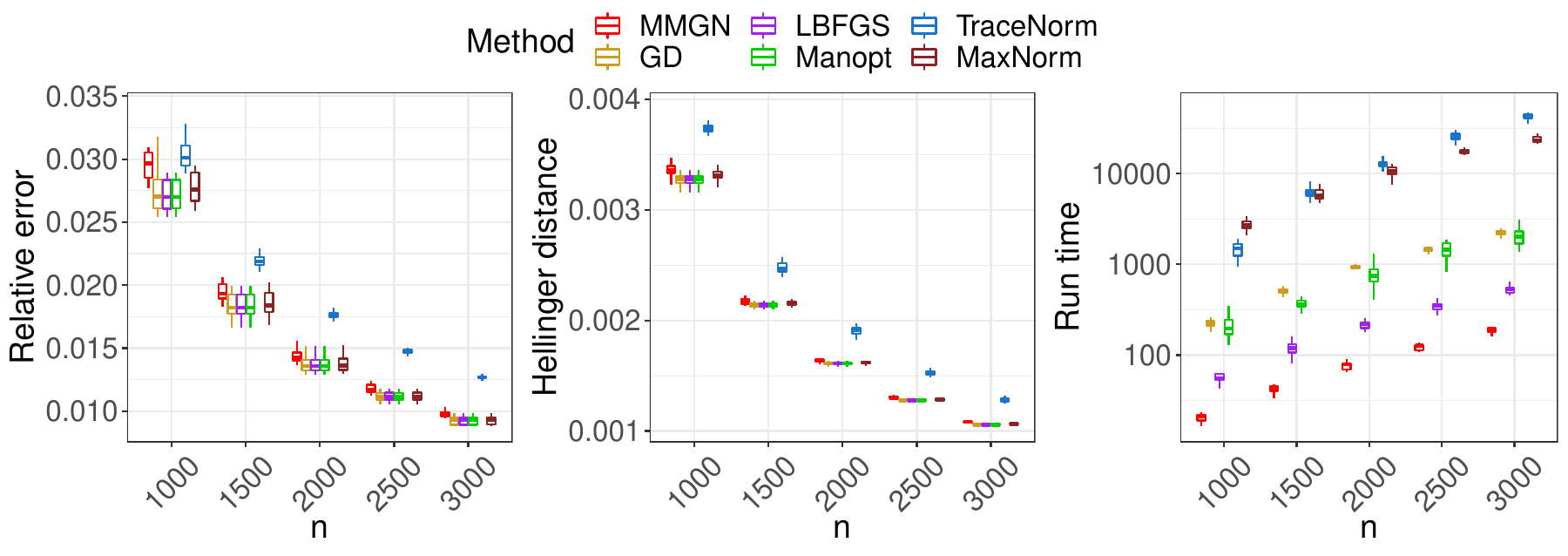}
    \caption{Probit model ($\sigma = 0.18$): Relative error, Hellinger distance, and runtime versus matrix dimension $n$ for a square non-spiky underlying matrix of rank $r^* = 5$ with observation fraction $\rho = 0.8$. }
    \label{fig:nonspiky-probit-dim-r1}
\end{figure}

\subsubsection{Varying the matrix rank}

In the last simulation under the non-spiky case, we examined how the performance of each method scaled with the rank of the underlying matrix. We fixed the size of $\M \Theta^*$ at  $m=n=1000$ and generated matrices of rank  $r^* \in \{3,5,8,10\}$.
The average spikiness ratios of the generated underlying matrices for the corresponding values of $r^*$ were $\{ 4.54, 4.81, 5.03, 4.97\}$.
We considered the probit model with a noise level $\sigma = 0.18$ and set the observation fraction $\rho = 0.8$. For all methods except for \texttt{MaxNorm}, we estimated the rank $r$ from $\{1, 2, \cdots, 12\}$. 


\begin{figure}[t]
    \centering \includegraphics[width=\textwidth]{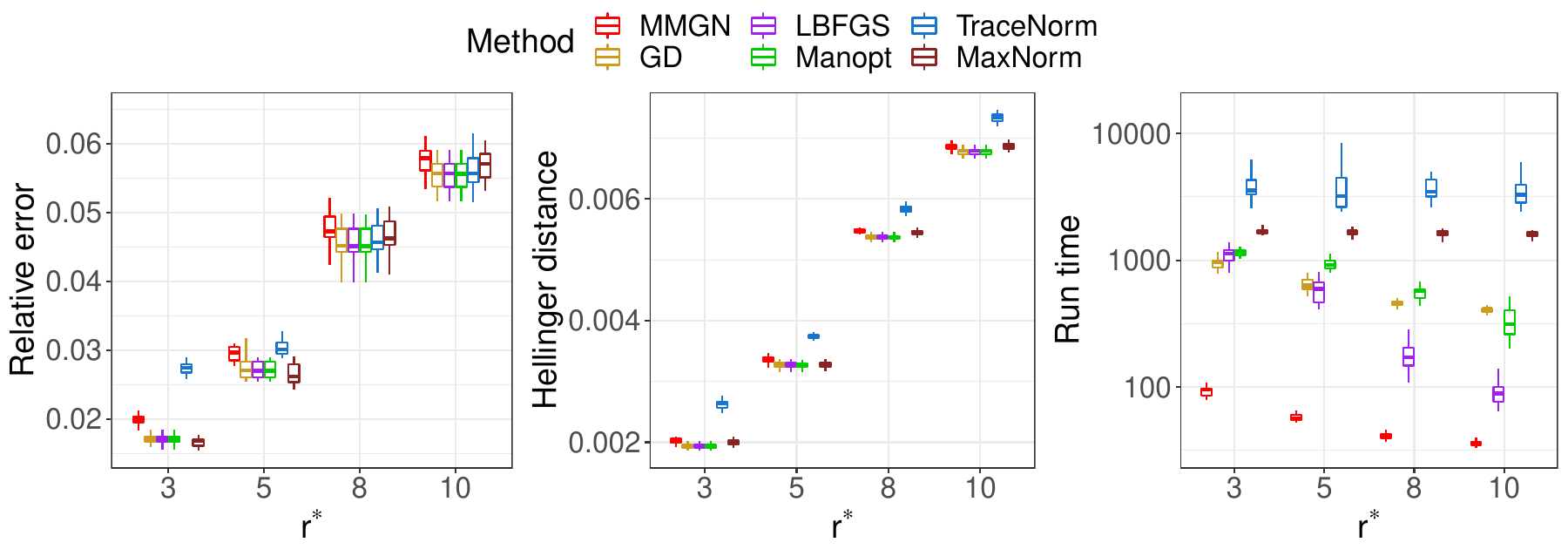}
    \caption{Probit model ($\sigma = 0.18$): Relative error, Hellinger distance, and runtime versus true rank $r^*$ for a non-spiky underlying matrix of size $m \times n = 1000 \times 1000$ and observation fraction $\rho = 0.8$. }
    \label{fig:nonspiky-probit-rSeq}
\end{figure}


\begin{figure}[t]
    \centering \includegraphics[width=\textwidth]{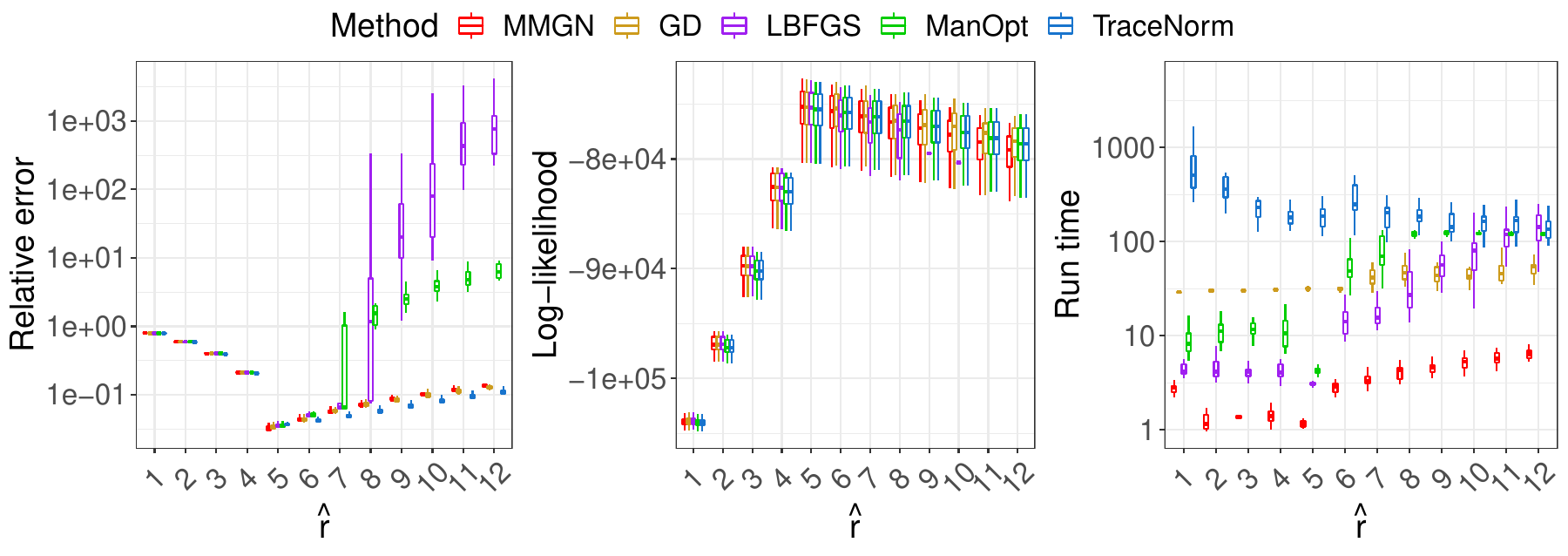}
    \caption{Probit model ($\sigma = 0.18$): Relative error (using the training set), log-likelihood (on the testing set), and runtime versus estimated rank $\hat{r}$ for a non-spiky underlying matrix of size $m \times n = 1000 \times 1000$, true rank $r^*=5$, and observation fraction $\rho = 0.8$. \texttt{LBFGS} produced log-likelihood of negative infinity at $\hat{r}\in \{11, 12\}$, so the corresponding box plots were missing.}
    \label{fig:nonspiky-rcheck}
\end{figure}

\Fig{nonspiky-probit-rSeq} displays the performance of different methods under varying rank $r^*$. As expected,  as $r^*$ increased, the estimation error of each method increased since the number of parameters to be estimated increased. All methods behaved comparably in estimation accuracy, with the Hellinger distance of \texttt{TraceNorm} slightly worse than the other methods across different ranks.
\texttt{MMGN} enjoyed the fastest runtime among all the compared methods. Recall that we provided the true rank $r^*$ for  \texttt{MaxNorm}, thus its runtime would be even longer if the rank were selected by validation.

To evaluate the rank estimation procedure and illustrate how different methods perform given different estimated ranks, we reran the above simulation with the true rank $r^*=5$ and reported the relative error using the training set, the log-likelihood on the testing set, and the runtime of each method when given an estimated rank $\hat{r}$ from the set $\{1, 2, \cdots, 12\}$. Note that \texttt{MaxNorm} was not included since we ran it without the rank estimation procedure.  As shown in \Fig{nonspiky-rcheck}, all methods achieved the highest log-likelihood as well as the lowest estimation error at the true rank $r^*=5$. In other words, they selected the correct rank by the procedure. As shown in the right panel of \Fig{nonspiky-rcheck}, \texttt{MMGN}, \texttt{GD}, \texttt{LBFGS}, and \texttt{Manopt} required longer runtimes when given a rank higher than the true rank. This explains why the runtimes of the four methods in \Fig{nonspiky-probit-rSeq} decreased as the true rank $r^*$ increased. With a fixed set of $\hat{r} \in \{1, 2, \cdots, 12\}$, as $r^*$ increased from $3$ to $10$, the number of values of $\hat{r}$ that were higher than $r^*$ decreased, which led to shorter overall runtime.

\subsection{Spiky Matrices}
\label{simulation-spiky}

We now consider 1-bit matrix completion with spiky underlying matrices, which has not been investigated in the literature. We generated spiky matrices as described in Section~\ref{sec: 3.1}. Specifically, we considered a square matrix $\M{\Theta}^* \in \Real^{m \times n}$ with $m=n=1000$ and rank $r^*=1$. 
We set the degrees of freedom $\nu$ to $10$, $5$, and $4$ to generate underlying matrices at low, intermediate, and high spikiness levels. The corresponding average spikiness ratios (with standard deviation in the parenthesis) over $20$ replicates are $17.57~ (2.02)$, $33.16~  (4.94)$, and $47.19~(5.65)$, respectively.
We considered the probit model with a noise level $\sigma = 2$. At each spikiness level, we varied the observation fraction $\rho$ over $\{0.2, 0.3, \cdots, 1\}$. We estimated the rank $r$ from $\{1,2,3,4, 5\}$ using the validation procedure for \texttt{MMGN}, \texttt{TraceNorm}, \texttt{GD}, \texttt{LBFGS} and \texttt{Manopt} but provided the true rank for \texttt{MaxNorm}.


\begin{figure}
    \centering    \includegraphics[width=\textwidth]{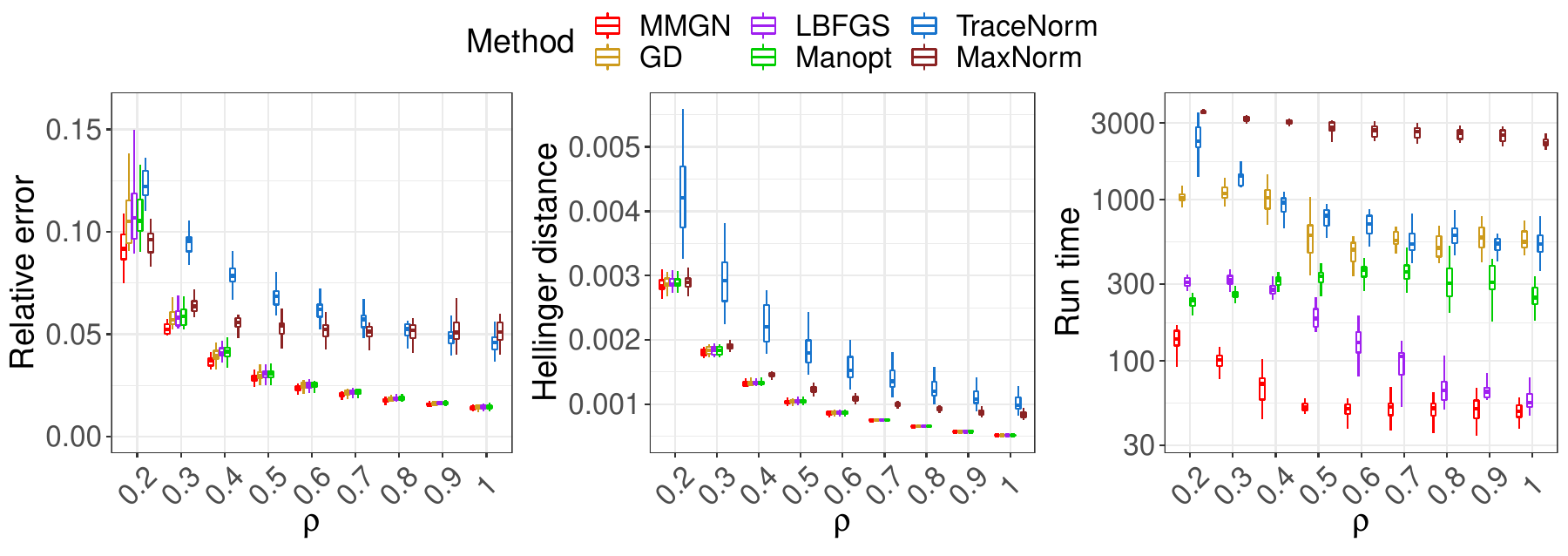}\\
    \includegraphics[width=\textwidth]{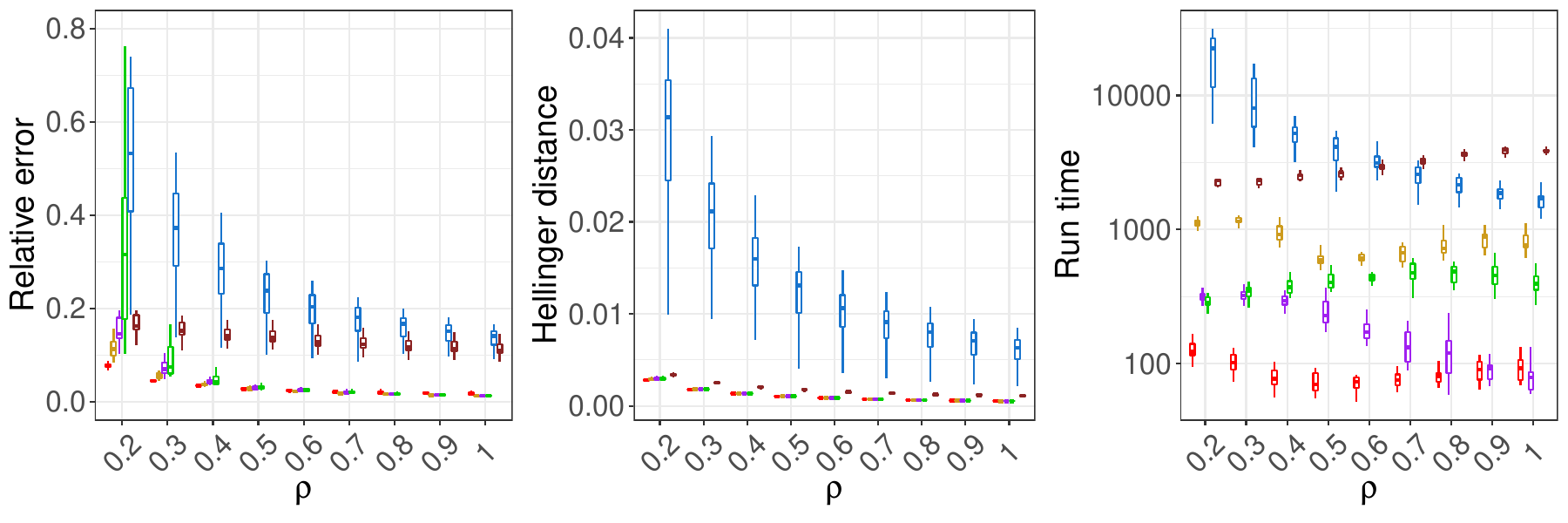}\\
    \includegraphics[width=\textwidth]{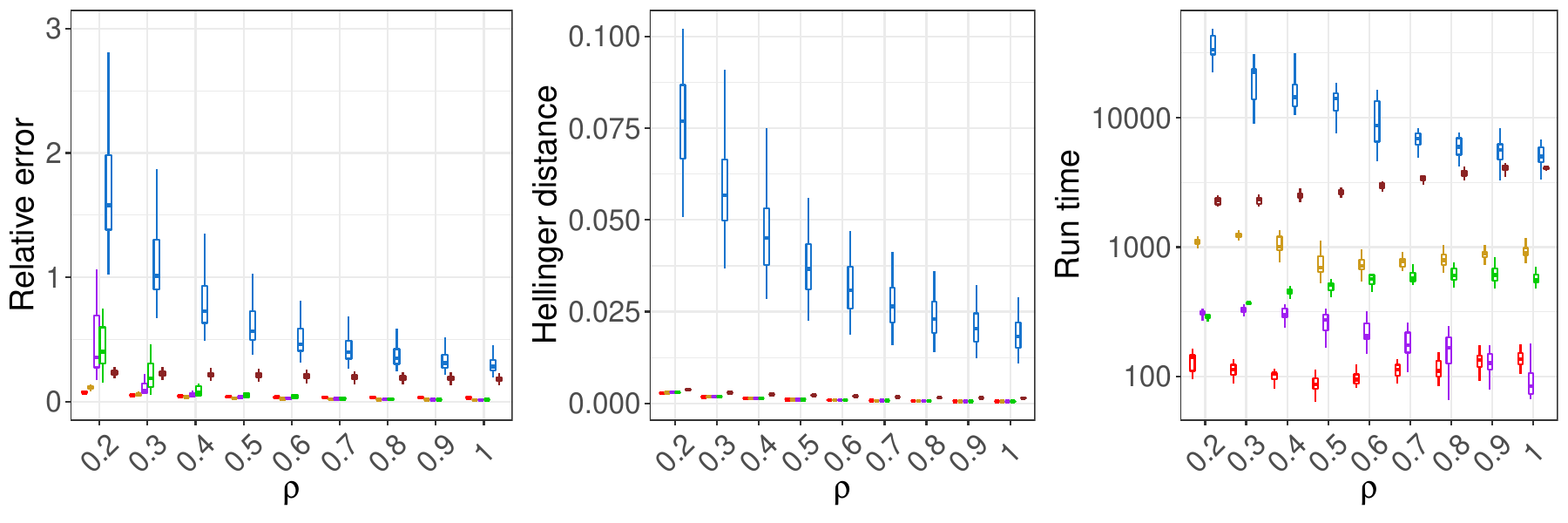}\\
    \caption{Probit model ($\sigma = 2$): Relative error, Hellinger distance, and runtime versus observation fraction $\rho$ at low (top), intermediate (middle), and high (bottom) spikiness levels for an underlying matrix of size $m \times n = 1000 \times 1000$ and rank $r^*=1$. Average spikiness ratios are $17.57$ (top), $33.16$ (middle), and $47.19$ (bottom).}
    \label{fig:spiky-t}
\end{figure}

\Fig{spiky-t} shows  
 results under the three spikiness levels. The top panel shows that at the low spikiness level, 
\texttt{MMGN} and the three generic schemes consistently achieved higher accuracy, in recovering both the underlying matrix $\M \Theta^*$ and the distribution matrix $\Phi(\M \Theta^*)$, than \texttt{TraceNorm} and \texttt{MaxNorm} over a wide range of values of $\rho$. \texttt{MaxNorm} achieved smaller errors in comparison to \texttt{TraceNorm} 
when $\rho < 0.8$ but larger errors at higher observation fractions when $\rho \geq 0.8$. However, \texttt{MaxNorm} recovered the underlying distribution more accurately than \texttt{TraceNorm}, as shown in the top middle panel of \Fig{spiky-t}. 

As the spikiness level grew, all methods suffered from lower estimation accuracy, but \texttt{MMGN} and the three generic schemes performed better than \texttt{TraceNorm} and \texttt{MaxNorm} in estimating $\M \Theta^*$. \texttt{MaxNorm}, \texttt{MMGN}, and the three generic schemes achieved comparable Hellinger distances and significantly outperformed \texttt{TraceNorm} at intermediate and high spikiness levels. Among all compared methods,  \texttt{TraceNorm} was the most sensitive to the spikiness level.

An interesting phenomenon is that for all methods except  \texttt{TraceNorm}, their errors in estimating $\M \Theta^*$ became relatively insensitive to $\rho$ as the spikiness level increased.  The relative errors were dominated by errors in the large-magnitude entries, and errors in these entries were mildly affected by $\rho$. These methods produced accurate estimates for the majority of low-magnitude entries, even when $\rho$ was small. Therefore, increasing  $\rho$ did not lead to significant improvement in their estimation errors. In contrast,  \texttt{TraceNorm} required a large value of $\rho$ to produce good estimates for low-magnitude entries. As a result, we saw a more noticeable decrease in its estimation error as $\rho$ grew.  We present a detailed example to elaborate on this phenomenon in the supplementary materials.

The right panel of each row in \Fig{spiky-t} shows the computational advantage of \texttt{MMGN} when dealing with spiky matrices at different spikiness levels. It ran dozens of times faster than \texttt{TraceNorm} and \texttt{MaxNorm}. Compared to the three generic schemes for the optimization problem \eqref{eq:1-bit-opt}, \texttt{MMGN} ran typically two to ten times faster. \texttt{LBFGS} exhibited runtimes comparable with \texttt{MMGN} but only when the observation fraction $\rho$ was close to one.

The simulations so far investigated metrics of overall accuracy defined as the average of the squared errors over all entries of the matrix. 
However, a more careful inspection reveals that the methods perform differently depending on the magnitude of entries of the underlying matrix $\M{\Theta}^*$. To better understand this more fine-grained behavior, we studied the errors 
conditioned on the underlying values $\theta_{ij}^*$, for a matrix $\M \Theta^*$ with a low spikiness ratio.

Specifically, we considered an underlying matrix $\M \Theta^*$ of size $m \times n =1000 \times 1000$, rank $r^* = 1$, and spikiness ratio $s=19.55$. The observation fraction $\rho = 0.8$. 
The resulting overall relative errors for \texttt{MMGN}, \texttt{TraceNorm}, and \texttt{MaxNorm} were $1.84 \times 10^{-2}$, $4.93 \times 10^{-2}$, and $5.6 \times 10^{-2}$. The corresponding Hellinger distances were $6.30\times 10^{-4}$, $1.10\times 10^{-3}$, and $9.55 \times 10^{-4}$, respectively. We divided entries of $\M{\Theta}^*$ into Value-Ranges that $\theta_{ij}^*$ took on. These Value-Ranges have corresponding Probability-Ranges. For instance, Value-Range $(-\infty, -2.5]$ corresponds to Probability-Range $(0, 0.1]$ since $\ \lim_{\theta \to \infty }\Phi(\theta) = 0$ and $\Phi(-2.5)\approx 0.1$ under the considered probit model with $\sigma = 2$.

\begin{figure}[t]
    \centering    \includegraphics[width=\textwidth]{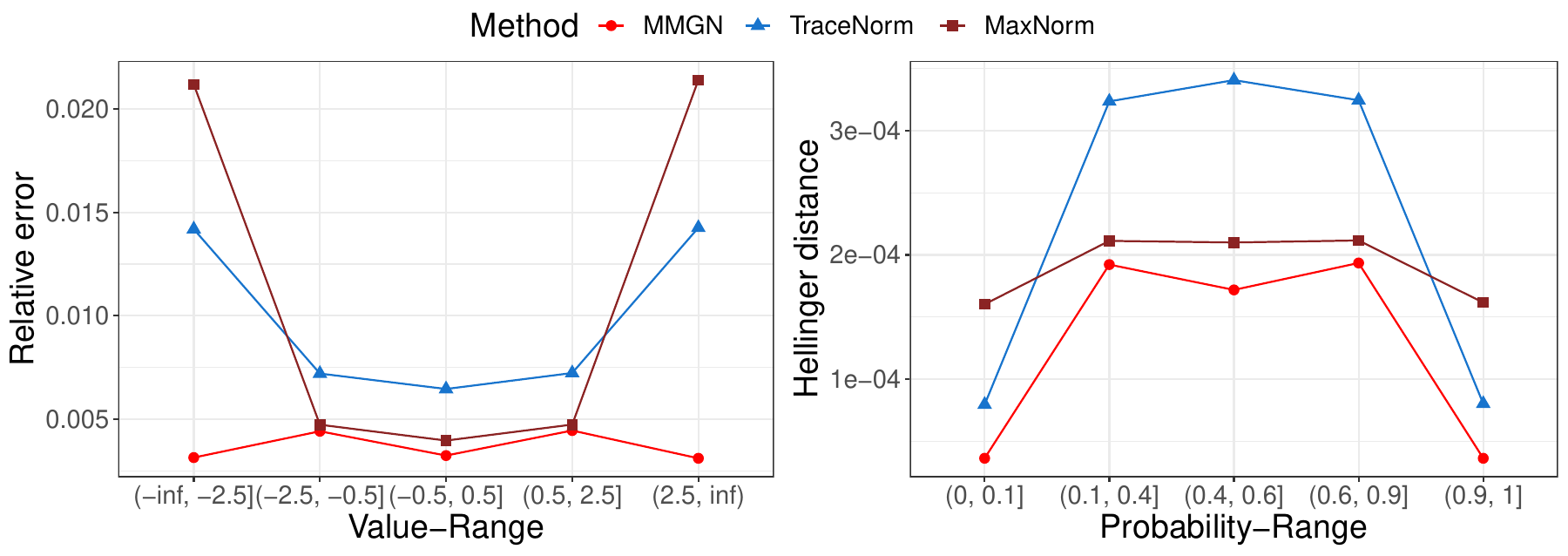}
    \caption{Estimation performance of different methods versus underlying matrix entry values $\ME{\theta}{ij}$ and probabilities $\Phi(\ME{\theta}{ij})$ for a spiky underlying matrix of size $m \times n =1000 \times 1000$, rank $r^* = 1$, and spikiness ratio $s=19.55$. 
    }
    \label{fig:spiky-details}
\end{figure}

\Fig{spiky-details} shows the performance of each method over the different Value- and Probability-Ranges. \texttt{TraceNorm} and \texttt{MaxNorm} yielded accurate estimates for small-magnitude entries of $\M \Theta^*$ (absolute values less than or equal to $2.5$) but poorer estimates for large-magnitude entries. 
\texttt{MaxNorm} suffered from much larger errors in estimating large-magnitude entries, leading to an overall relative error larger than \texttt{TraceNorm}. However, \texttt{MaxNorm} accurately estimated small-magnitude entries leading to accurate recoveries of the corresponding $\Phi(\theta^*_{ij})$ and a smaller Hellinger distance. This discrepancy is caused by the nonlinear transformation from $\theta_{ij}^*$ to $\Phi(\theta_{ij}^*)$. For example, $\Phi(0.5)- \Phi(0.4) \approx 0.019$, while $\Phi(3.6) - \Phi(3.5) \approx 0.004$. In other words, the recovered distribution is more sensitive to the accuracy of the recovered $\hat{\theta}_{ij}$ in small-magnitude groups. In comparison, \texttt{MMGN} achieved the highest accuracy across different Value-Ranges, leading to the best overall performance. \Fig{spiky-details} provides a more detailed picture of how \texttt{MMGN} outperformed \texttt{TraceNorm} and \texttt{MaxNorm}. Moreover, this explains why the trends in estimating $\M{\Theta}^*$ do not mirror the trends in estimating $\Phi(\M \Theta^*)$ seen in Figures~\ref{fig:nonspiky-noise} and \ref{fig:spiky-t}.

\subsection{Real Data Example}
\label{sec: real-data}

We applied \texttt{MMGN} to the MovieLens (1M) data set\footnote{The data set is available at \href{http://www.grouplens.org/node/73}{http://www.grouplens.org/node/73}.}, which is larger but otherwise similar to the MovieLens (100K) real data example in \cite{davenport20141}. 
It contains
$1{,}000{,}209$ movie ratings from $6{,}040$ users on $3{,}952$ movies, with each
rating on a scale from $1$ to $5$. The resulting observation fraction is $0.04$. Following \cite{davenport20141}, we converted all ratings to binary observations by comparing each rating
to the average rating over all movies. Ratings above the average were encoded as $+1$ and $-1$ otherwise. 
As in \cite{davenport20141}, we considered the logistic model with the noise level $\sigma=1$ and used $95\%$ of the ratings as the training set to estimate $\M{\Theta}^*$. The performance was evaluated on the remaining $5\%$ of ratings by checking whether or not the estimate of $\M{\Theta}^*$ correctly predicts the
sign of the ratings (above or below the average rating).
 We compared \texttt{MMGN} to \texttt{TraceNorm},  \texttt{MaxNorm}, \texttt{GD}, \texttt{LBFGS}, and \texttt{Manopt}. We followed \cite{davenport20141} and treated $\alpha \beta$ as a single parameter in \texttt{TraceNorm}. We assigned $\alpha \beta$ ten values equally spaced on a logarithmic scale between $10^{-0.5}$ to $10$. 
 We ran \texttt{MMGN} and the three generic schemes with a candidate rank $r$ ranging from $1$ to $10$.  
For \texttt{MaxNorm}, $\beta$ was set to $\sqrt{r}$, where again $r$ ranged from $1$ to $10$, and $\alpha$ was chosen such that $\alpha\beta$ took the same value as $\alpha\beta$ for  \texttt{TraceNorm}.
 As in \cite{davenport20141}, we selected parameter values for each method that led to the best prediction performance on the separate validation set.


\Tab{Table-real-data} summarizes the median prediction accuracies and runtimes of the compared methods over $20$ replicates. 
All methods achieved comparable prediction accuracies, but \texttt{MMGN} was more than an order of magnitude faster than the two existing 1-bit matrix completion methods and at least five-fold faster than the three generic schemes. This example highlights the computational advantage of \texttt{MMGN} in dealing with large sparse (very few observations) data sets from real-world problems.


\begin{table}
\def~{\hphantom{0}}
{%
\caption{Performance of different methods on  MovieLens (1M) data}
\label{tab:Table-real-data}
\begin{tabular}{lcccccc}
\\
Method &  \texttt{MMGN} & \texttt{GD} & \texttt{LBFGS} &  \texttt{Manopt} &  \texttt{TraceNorm} & \texttt{MaxNorm}  \\ \hline 
 Accuracy (\%) & 74.2 & 74.4 & 74.0 & 74.1 & 75.0 & 74.0  \\
 Time (seconds) & 6.7e+2  & 3.36e+3 & 2.25e+04 & 4.08e+03 &  1.20e+4 &  6.41e+4\\
\hline 
\end{tabular}}
\end{table}

\subsection{Additional Comparisons}

As mentioned above, 1-bit matrix completion can be viewed as 
a logistic PCA problem but with only partially observed entries. In particular, one could solve the 1-bit objective by a logistic-PCA scheme that is able to handle missing data. \cite{de2006principal} proposed such a scheme, denoted \texttt{logisticPCA}. Thus, we compared \texttt{MMGM} to \texttt{logisticPCA}. 
Interestingly, this scheme is also based on MM. The difference between the two algorithms lies in how they solve the rank-constrained least squares problem. The per-iteration computation in {\texttt{logisticPCA} is dominated by a rank-$r$ truncated SVD. The implementation we used for \texttt{logisticPCA} is based on ARPACK which requires $r$ rounds of the Lanczos method for a total cost of $O(mnr)$ flops; each round of the Lanczos method costs $O(mn)$ flops, due to matrix-vector multiplies involving an $m$-by-$n$ matrix \citep{ARPACK}. In contrast, \texttt{MMGN} adopts a single Gauss-Newton step to approximate the rank-constrained least squares problem by a sparse least square problem whose computational cost is $\mathcal{O}(|\Omega|r)$. A detailed breakdown of this cost is provided in the supplementary materials. 
In simulations, which can be found in the supplementary materials, \texttt{logisticPCA} typically achieved accuracies comparable with  \texttt{MMGN} but ran two to ten times slower than \texttt{MMGN}, especially in the more challenging cases where the observation fraction is small.

We also compared \texttt{MMGN} to the nonnegative binary matrix factorization (\texttt{NBMF-MM}) method proposed by \cite{magron2022majorization}. \texttt{NBMF-MM} is the state-of-the-art method for BMF, displaying computational advantages over competing BMF methods \citep{kaban2008factorisation, bingham2009aspect, lumbreras2020bayesian}, as discussed in \cite{magron2022majorization}. Given a target rank of $r$,
the per-iteration computational cost of {\texttt{NBMF-MM} is $\mathcal{O}(mnr)$ flops. In simulations, which can be found in the supplementary materials, \texttt{MMGN} was more accurate than \texttt{NBMF-MM}, while running typically dozens of times faster.

We also compared \texttt{MMGN} with a 1-bit tensor completion (\texttt{1BitTC}) method \citep{binary-tensor} viewing our matrix as a three-way tensor with the third mode having dimension one. While several methods were proposed for binary tensor completion \citep{li2018tensor, aidini2018,ghadermarzy2018learning, binary-tensor}, we focused on \texttt{1BitTC} since, to the best of our knowledge, it has the best error rates, as well as publicly available code. Since \texttt{1BitTC} works with tensors, it is expected to be much slower than methods that work directly with matrices. 
As shown in the supplementary materials, in addition to being slower, \texttt{1BitTC} produced higher errors under many scenarios compared with  \texttt{MMGN}.

\section{Summary and Discussion}
\label{sec: discussion}


In this paper, we proposed 
\texttt{MMGN},
a novel fast and accurate algorithm for solving the 1-bit optimization problem.} Our algorithm employs the MM principle to convert the original 
challenging problem \eqref{eq:1-bit-opt} into a sequence of standard low-rank matrix completion problems. For each MM update, we apply a factorization strategy to incorporate the low-rank structure, resulting in a nonlinear least squares problem. We then apply a modified Gauss-Newton scheme to compute an inexact MM update by solving a linear least squares problem. 
Hence, in comparison to previous works, our method is relatively simple and can easily scale to large matrices. In comparison to two established 1-bit matrix completion methods, \texttt{TraceNorm}  \citep{davenport20141} and \texttt{MaxNorm}  \citep{cai2013max}, and three generic schemes for solving the original optimization problem \eqref{eq:1-bit-opt}, \texttt{MMGN} typically achieved comparable and sometimes lower errors while often being significantly faster.

Our work raises several interesting directions for further research. On the practical side, it is of interest to extend \texttt{MMGN} to other quantization models, such as movie ratings between 1 and 5. On the theoretical side, an open problem is to prove that under suitable assumptions, and possibly starting from a sufficiently accurate initial guess, \texttt{MMGN} indeed converges to the maximum likelihood solution. 

\section*{Supplementary Materials}
\begin{description}

\item[Title:] Supplementary Materials for ``A Majorization-Minimization Gauss-Newton Method for 1-Bit Matrix Completion". (.pdf file)



\end{description}

\section*{Acknowledgments}
B.N. is incumbent of the William Petschek
Professorial Chair of Mathematics. The research of B.N. 
was funded by the  National Institutes of Health (NIH) grant R01GM135928 and by the Israel Science Foundation (ISF) grant 2362/22. The research of E.C. was  supported by the NIH grant R01GM135928. We thank Wenxin Zhou for sharing us the code of \texttt{MaxNorm}. We thank Mark Davenport for helpful discussions. 

\bibliographystyle{asa}

\bibliography{main}

\begin{thebibliography}{37}
\newcommand{\enquote}[1]{``#1''}
\expandafter\ifx\csname natexlab\endcsname\relax\def\natexlab#1{#1}\fi

\bibitem[{Aidini et~al.(2018)Aidini, Tsagkatakis, and Tsakalides}]{aidini2018}
Aidini, A., Tsagkatakis, G., and Tsakalides, P. (2018), \enquote{1-bit tensor completion,} \textit{Electronic Imaging}, 30, 261--1-- 261--6.

\bibitem[{Bauch et~al.(2021)Bauch, Nadler, and Zilber}]{bauch2021rank}
Bauch, J., Nadler, B., and Zilber, P. (2021), \enquote{Rank 2r iterative least squares: Efficient recovery of ill-conditioned low rank matrices from few entries,} \textit{SIAM Journal on Mathematics of Data Science}, 3, 439--465.

\bibitem[{Bhaskar and Javanmard(2015)}]{bhaskar20151}
Bhaskar, S.~A. and Javanmard, A. (2015), \enquote{1-bit matrix completion under exact low-rank constraint,} in \textit{2015 49th Annual Conference on Information Sciences and Systems (CISS)}, IEEE, pp. 1--6.

\bibitem[{Bingham et~al.(2009)Bingham, Kab{\'a}n, and Fortelius}]{bingham2009aspect}
Bingham, E., Kab{\'a}n, A., and Fortelius, M. (2009), \enquote{The aspect Bernoulli model: multiple causes of presences and absences,} \textit{Pattern Analysis and Applications}, 12, 55--78.

\bibitem[{Boumal et~al.(2014)Boumal, Mishra, Absil, and Sepulchre}]{manopt}
Boumal, N., Mishra, B., Absil, P.-A., and Sepulchre, R. (2014), \enquote{{M}anopt, a {M}atlab Toolbox for Optimization on Manifolds,} \textit{Journal of Machine Learning Research}, 15, 1455--1459.

\bibitem[{Cai and Zhou(2013)}]{cai2013max}
Cai, T. and Zhou, W.-X. (2013), \enquote{A max-norm constrained minimization approach to 1-bit matrix completion,} \textit{The Journal of Machine Learning Research}, 14, 3619--3647.

\bibitem[{Collins et~al.(2001)Collins, Dasgupta, and Schapire}]{collins2001generalization}
Collins, M., Dasgupta, S., and Schapire, R.~E. (2001), \enquote{A generalization of principal components analysis to the exponential family,} \textit{Advances in Neural Information Processing Systems}, 14.

\bibitem[{Davenport et~al.(2014)Davenport, Plan, Van Den~Berg, and Wootters}]{davenport20141}
Davenport, M.~A., Plan, Y., Van Den~Berg, E., and Wootters, M. (2014), \enquote{1-bit matrix completion,} \textit{Information and Inference: A Journal of the IMA}, 3, 189--223.

\bibitem[{De~Leeuw(1994)}]{deLeeuw1994}
De~Leeuw, J. (1994), \enquote{Block-relaxation algorithms in statistics,} in \textit{Information Systems and Data Analysis}, Springer Berlin Heidelberg, pp. 308--324.

\bibitem[{De~Leeuw(2006)}]{de2006principal}
--- (2006), \enquote{Principal component analysis of binary data by iterated singular value decomposition,} \textit{Computational Statistics \& Data Analysis}, 50, 21--39.

\bibitem[{Ghadermarzy et~al.(2018)Ghadermarzy, Plan, and Yilmaz}]{ghadermarzy2018learning}
Ghadermarzy, N., Plan, Y., and Yilmaz, O. (2018), \enquote{Learning tensors from partial binary measurements,} \textit{IEEE Transactions on Signal Processing}, 67, 29--40.

\bibitem[{Gross et~al.(2010)Gross, Liu, Flammia, Becker, and Eisert}]{quantum}
Gross, D., Liu, Y.-K., Flammia, S.~T., Becker, S., and Eisert, J. (2010), \enquote{Quantum state tomography via compressed sensing,} \textit{Physical Review Letters}, 105, 150401.

\bibitem[{Heiser(1995)}]{Heiser1995}
Heiser, W.~J. (1995), \enquote{Convergent computation by iterative majorization,} \textit{Recent Advances in Descriptive Multivariate Analysis}, 157--189.

\bibitem[{Hestenes et~al.(1952)Hestenes, Stiefel, et~al.}]{hestenes1952methods}
Hestenes, M.~R., Stiefel, E., et~al. (1952), \enquote{Methods of conjugate gradients for solving linear systems,} \textit{Journal of Research of the National Bureau of Standards}, 49, 409--436.

\bibitem[{Hunter and Lange(2004)}]{HunterLange2004}
Hunter, D.~R. and Lange, K. (2004), \enquote{A tutorial on MM algorithms,} \textit{The American Statistician}, 58, 30--37.

\bibitem[{Kab{\'a}n and Bingham(2008)}]{kaban2008factorisation}
Kab{\'a}n, A. and Bingham, E. (2008), \enquote{Factorisation and denoising of 0--1 data: a variational approach,} \textit{Neurocomputing}, 71, 2291--2308.

\bibitem[{Kammerer and Nashed(1972)}]{kammerer1972convergence}
Kammerer, W.~J. and Nashed, M.~Z. (1972), \enquote{On the convergence of the conjugate gradient method for singular linear operator equations,} \textit{SIAM Journal on Numerical Analysis}, 9, 165--181.

\bibitem[{Koren et~al.(2009)Koren, Bell, and Volinsky}]{koren2009matrix}
Koren, Y., Bell, R., and Volinsky, C. (2009), \enquote{Matrix factorization techniques for recommender systems,} \textit{Computer}, 42, 30--37.

\bibitem[{Lange(2016)}]{lange2016mm}
Lange, K. (2016), \textit{MM Optimization Algorithms}, Philadelphia, PA, USA: SIAM.

\bibitem[{Lange et~al.(2000)Lange, Hunter, and Yang}]{lange2000optimization}
Lange, K., Hunter, D.~R., and Yang, I. (2000), \enquote{Optimization transfer using surrogate objective functions,} \textit{Journal of Computational and Graphical Statistics}, 9, 1--20.

\bibitem[{Lehoucq et~al.(1998)Lehoucq, Sorensen, and Yang}]{ARPACK}
Lehoucq, R.~B., Sorensen, D.~C., and Yang, C. (1998), \textit{ARPACK users' guide - solution of large-scale eigenvalue problems with implicitly restarted Arnoldi methods.}, SIAM.

\bibitem[{Li et~al.(2018)Li, Zhang, Li, and Lu}]{li2018tensor}
Li, B., Zhang, X., Li, X., and Lu, H. (2018), \enquote{Tensor completion from one-bit observations,} \textit{IEEE Transactions on Image Processing}, 28, 170--180.

\bibitem[{Liben-Nowell and Kleinberg(2003)}]{liben2003link}
Liben-Nowell, D. and Kleinberg, J. (2003), \enquote{The link prediction problem for social networks,} in \textit{Proceedings of the Twelfth International Conference on Information and Knowledge Management}, pp. 556--559.

\bibitem[{Linial et~al.(2007)Linial, Mendelson, Schechtman, and Shraibman}]{linial2007complexity}
Linial, N., Mendelson, S., Schechtman, G., and Shraibman, A. (2007), \enquote{Complexity measures of sign matrices,} \textit{Combinatorica}, 27, 439--463.

\bibitem[{Lumbreras et~al.(2020)Lumbreras, Filstroff, and F{\'e}votte}]{lumbreras2020bayesian}
Lumbreras, A., Filstroff, L., and F{\'e}votte, C. (2020), \enquote{Bayesian mean-parameterized nonnegative binary matrix factorization,} \textit{Data Mining and Knowledge Discovery}, 34, 1898--1935.

\bibitem[{Magron and F{\'e}votte(2022)}]{magron2022majorization}
Magron, P. and F{\'e}votte, C. (2022), \enquote{A majorization-minimization algorithm for nonnegative binary matrix factorization,} \textit{IEEE Signal Processing Letters}, 29, 1526--1530.

\bibitem[{Miller et~al.(2009)Miller, Jordan, and Griffiths}]{miller2009nonparametric}
Miller, K., Jordan, M., and Griffiths, T. (2009), \enquote{Nonparametric latent feature models for link prediction,} \textit{Advances in Neural Information Processing Systems}, 22.

\bibitem[{Negahban and Wainwright(2012)}]{spikiness}
Negahban, S. and Wainwright, M.~J. (2012), \enquote{Restricted strong convexity and weighted matrix completion: Optimal bounds with noise,} \textit{The Journal of Machine Learning Research}, 13, 1665--1697.

\bibitem[{Ni and Gu(2016)}]{ni2016optimal}
Ni, R. and Gu, Q. (2016), \enquote{Optimal statistical and computational rates for one bit matrix completion,} in \textit{Artificial Intelligence and Statistics}, PMLR, vol.~51, pp. 426--434.

\bibitem[{Nocedal and Wright(2006)}]{NoceWrig06}
Nocedal, J. and Wright, S.~J. (2006), \textit{Numerical Optimization}, New York, NY, USA: Springer, 2nd ed.

\bibitem[{Paige and Saunders(1982)}]{PaigeSaunders1982}
Paige, C.~C. and Saunders, M.~A. (1982), \enquote{LSQR: An algorithm for sparse linear equations and sparse least squares,} \textit{ACM Transactions on Mathematical Software}, 8, 43--71.

\bibitem[{Schein et~al.(2003)Schein, Saul, and Ungar}]{schein2003generalized}
Schein, A.~I., Saul, L.~K., and Ungar, L.~H. (2003), \enquote{A generalized linear model for principal component analysis of binary data,} in \textit{International Workshop on Artificial Intelligence and Statistics}, PMLR, pp. 240--247.

\bibitem[{Wang and Li(2020)}]{binary-tensor}
Wang, M. and Li, L. (2020), \enquote{Learning from binary multiway data: Probabilistic tensor decomposition and its statistical optimality,} \textit{Journal of Machine Learning Research}, 21, 6146--6183.

\bibitem[{Xu et~al.(2018)Xu, Chi, Yang, and Lange}]{Xu2018}
Xu, J., Chi, E.~C., Yang, M., and Lange, K. (2018), \enquote{A majorization--minimization algorithm for split feasibility problems,} \textit{Computational Optimization and Applications}, 71, 795--828.

\bibitem[{Zhou(2015{\natexlab{a}})}]{zhou2015infinite}
Zhou, M. (2015{\natexlab{a}}), \enquote{Infinite edge partition models for overlapping community detection and link prediction,} in \textit{Artificial Intelligence and Statistics}, PMLR, pp. 1135--1143.

\bibitem[{Zhou(2015{\natexlab{b}})}]{zhou2015nonparametric}
--- (2015{\natexlab{b}}), \enquote{Nonparametric Bayesian matrix factorization for assortative networks,} in \textit{2015 23rd European Signal Processing Conference (EUSIPCO)}, IEEE, pp. 2776--2780.

\bibitem[{Zilber and Nadler(2022)}]{ZilberNadler2022}
Zilber, P. and Nadler, B. (2022), \enquote{GNMR: A provable one-line algorithm for low rank matrix recovery,} \textit{SIAM Journal on Mathematics of Data Science}, 4, 909--934.

\end{thebibliography}
\end{document}